\documentclass[lettersize,journal]{IEEEtran}
\usepackage{amsmath,amssymb,amsfonts}
\usepackage{algorithmic}
\usepackage{algorithm}
\usepackage{array, makecell}
\usepackage{multirow}
\usepackage[caption=false,font=normalsize,labelfont=sf,textfont=sf]{subfig}
\usepackage{textcomp}
\usepackage{stfloats}
\usepackage{url}
\usepackage{verbatim}
\usepackage{graphicx}
\usepackage{cite}
\usepackage{adjustbox}
\usepackage{tikz}
\graphicspath{ {figure/} }
\usepackage{yhlee-journal}
\usepackage{setspace}
\usepackage[font=normalsize]{caption}

\begin{document}

\title{CoVOR-SLAM: Cooperative SLAM using Visual Odometry and Ranges for Multi-Robot Systems}

\author{Young-Hee Lee$^{*}$, Chen Zhu$^{*}$, Thomas Wiedemann$^{*}$, Emanuel Staudinger$^{*}$, Siwei Zhang$^{*}$, Christoph G\"{u}nther$^{*,**}$ 
\thanks{$^{*}$The authors are with the Institute of Communications and Navigation, German Aerospace Center (DLR), Oberpfaffenhofen, Germany
        {\tt\small <first name>.<last name>@dlr.de}}
\thanks{$^{**}$The author is also with the Institute for Communications and Navigation, Technical University of Munich, Munich, Germany
        {\tt\small christoph.guenther@tum.de}}
}




\maketitle

\begin{abstract}
A swarm of robots has advantages over a single robot, since it can explore larger areas much faster and is more robust to single-point failures.
Accurate relative positioning is necessary to successfully carry out a collaborative mission without collisions. When Visual Simultaneous Localization and Mapping (VSLAM) is used to estimate the poses of each robot, inter-agent loop closing is widely applied to reduce the relative positioning errors. This technique can mitigate errors using the feature points commonly observed by different robots. However, it requires significant computing and communication capabilities to detect inter-agent loops, and to process the data transmitted by multiple agents. 
In this paper, we propose Collaborative SLAM using Visual Odometry and Range measurements (CoVOR-SLAM) to overcome this challenge. In the framework of CoVOR-SLAM, robots only need to exchange pose estimates, covariances (uncertainty) of the estimates, and range measurements between robots. Since CoVOR-SLAM does not require to associate visual features and map points observed by different agents, the computational and communication loads are significantly reduced. The required range measurements can be obtained using pilot signals of the communication system, without requiring complex additional infrastructure.
We tested CoVOR-SLAM using real images as well as real ultra-wideband-based ranges obtained with two rovers. In addition, CoVOR-SLAM is evaluated with a larger scale multi-agent setup exploiting public image datasets and ranges generated using a realistic simulation.
The results show that CoVOR-SLAM can accurately estimate the robots' poses, requiring much less computational power and communication capabilities than the inter-agent loop closing technique.
\end{abstract}


\section{Introduction}

In these days, multi-robot systems are widely studied to carry out tasks that are challenging for a single agent (e.g. monitoring indoor warehouses, rescue missions, and space exploration, etc.). 
Accurate relative pose (positions and attitude) estimation is one of the key capabilities to successfully conduct such missions. 
When Visual Simultaneous Localization and Mapping (VSLAM) \cite{engel2014lsd,mur2017orb,davison2007monoslam} is used for positioning, loop closure procedures have been generally applied to mitigate errors. To align the generated maps and the position of the robots, the inter-agent loop closing technique detects the features commonly observed by different robots, and merges the local maps of different robots using the associated features \cite{schmuck2019ccm, wen2020, riazuelo2014c2tam, mohanarajah2015cloud, cunningham2010ddf, cunningham2013ddf, cieslewski2018data}. Significant computational power is required for processing visual information to detect inter-agent loops. Substantial communication loads is also needed to exchange large amounts of visual measurements between robots. Furthermore, the motion of the robots is significantly constrained since they have to be closely located or revisit the areas observed by other robots for detecting an inter-agent loop. 

\begin{figure}[t!]
\centering
\captionsetup[subfloat]{labelfont=normalsize,textfont=normalsize}
\captionsetup[subfloat]{labelfont=normalsize,textfont=normalsize}
\subfloat[Space exploration (image credit: \cite{zhang2020self})]{\centering\includegraphics[width=.95\linewidth]{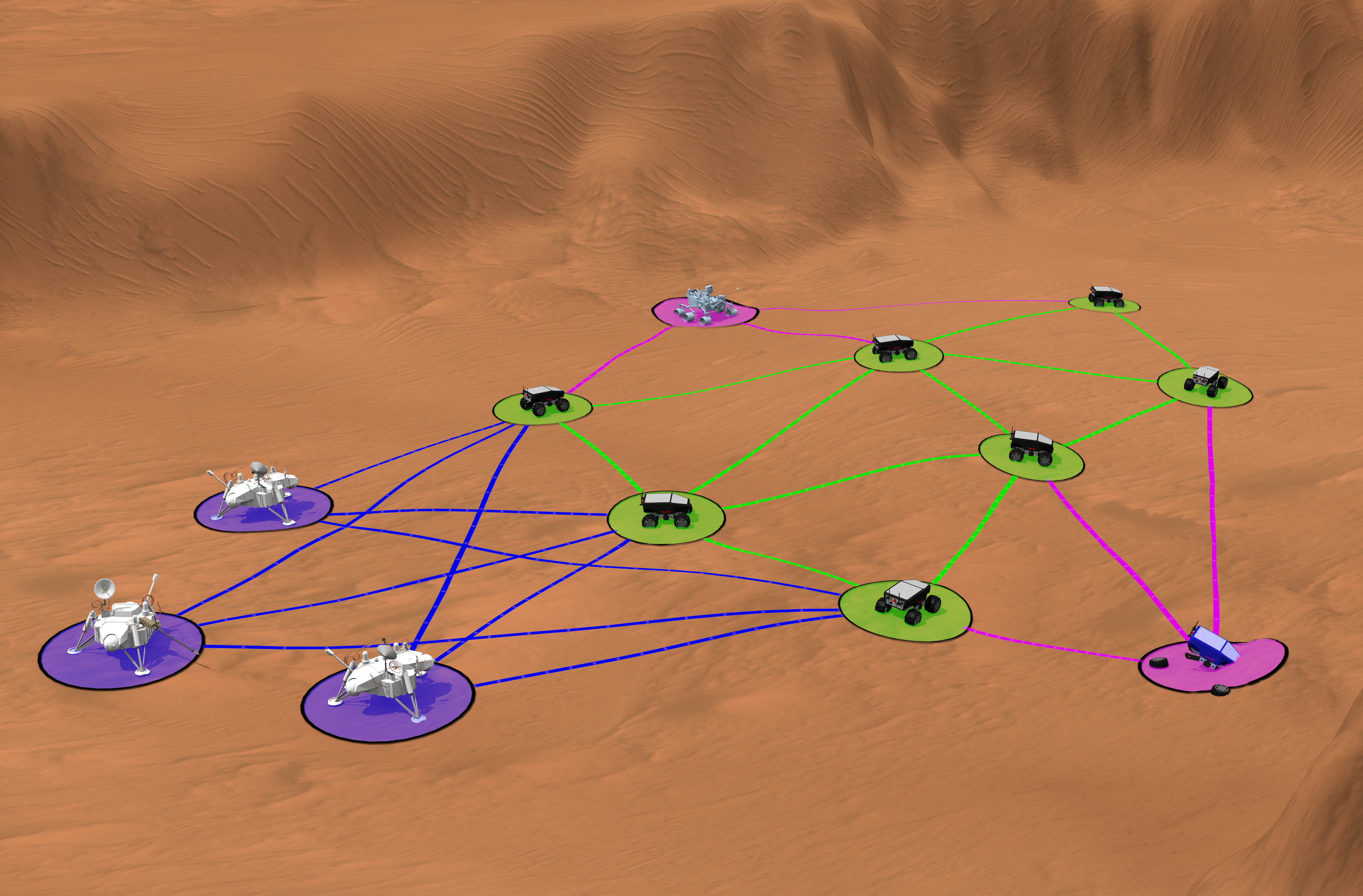}\label{subfig:intro_mars_exploration}}\\
\subfloat[Deep urban positioning using inter-agent communication links and a single anchor station in each cell (icon credits: \cite{treeByJino, buildingByIconsphere})]{\centering\includegraphics[width=.95\linewidth]{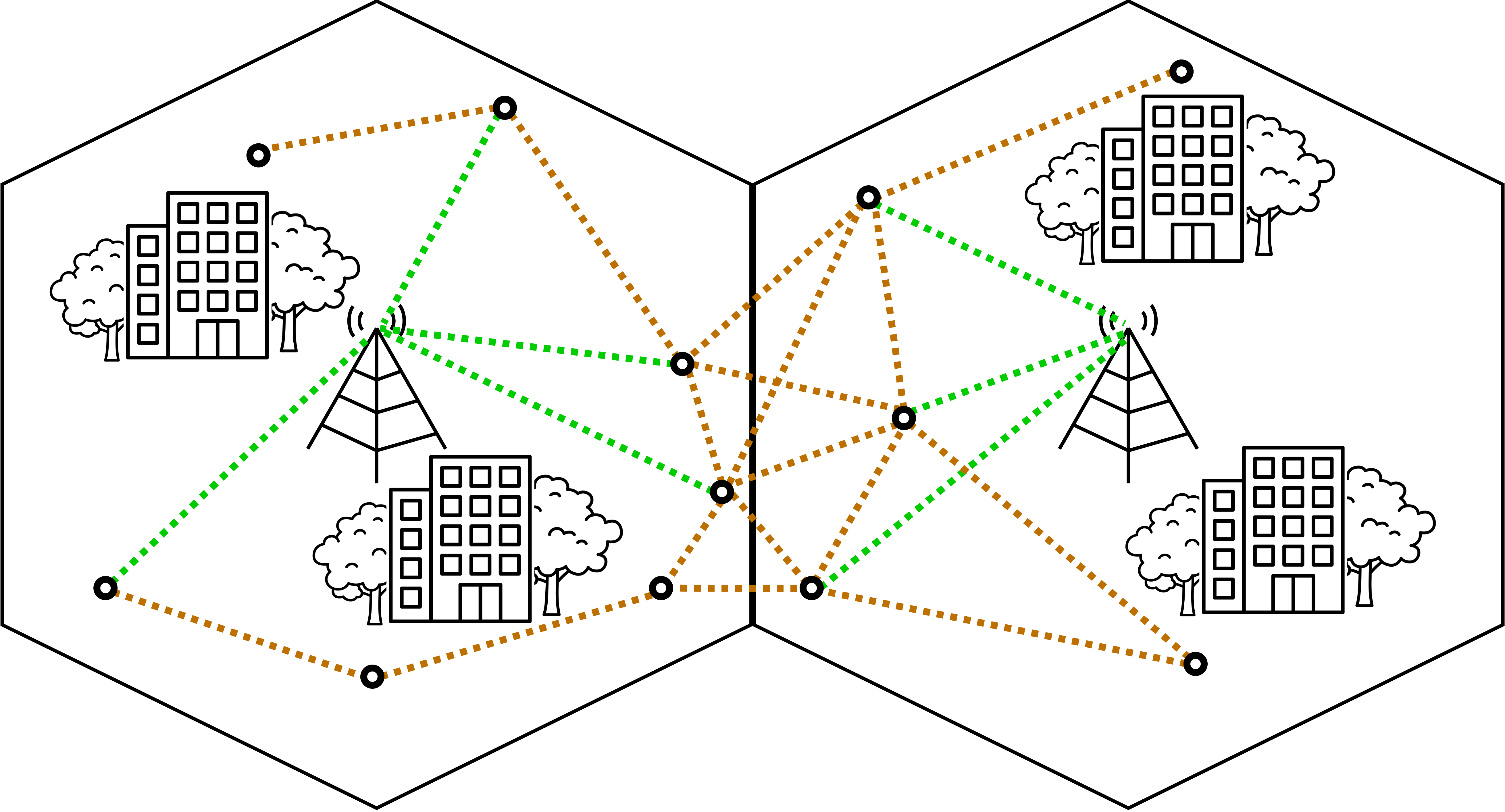}\label{subfig:intro_urban_applications}}
\caption{Application scenarios of CoVOR-SLAM with multi-robot systems}
\label{fig:intro_applications}
\end{figure}


To overcome the challenges of inter-agent loop closures, we propose \textbf{Co}llaborative \textbf{SLAM} using \textbf{V}isual \textbf{O}dometry and \textbf{R}anges (CoVOR-SLAM). For CoVOR-SLAM, vehicles only need to exchange their pose estimates, covariances (uncertainty) of their estimates, and inter-agent range measurements. 
Since robots do not need to transmit visual features, CoVOR-SLAM can be very useful when computational and communication capabilities are limited, e.g. in space exploration missions (see \figref{subfig:intro_mars_exploration}). 
The range measurements can be easily obtained from round-trip delay measurements using pilot signals of the communication systems of robots \cite{staudinger2014rtd}, enabling localization of the robots with limited number of anchor stations. Without the inter-robot ranging measurements, at least three or four anchor points (dependent on scenarios) are necessary to position the robots, which can be challenging in deep urban scenarios. Thus, CoVOR-SLAM can be useful in urban localization as shown in \figref{subfig:intro_urban_applications}, where line-of-sight connections to satellites and base stations are highly obstructed. 
As a relevant work in such use cases, Morales et al. proposed a collaborative SLAM method using inertial sensors and signals of opportunity as ranging sources in \cite{Morales2022CIRSLAM}.

CoVOR-SLAM is an extension of our previous works on cooperative SLAM using camera-ranging fusion. We first applied the concept in 2D scenarios \cite{zhu2018enhancing, zhu2019cooperative}, and then extended it to the 3D application cases in \cite{lee2018stereo,lee2018fusion,lee2020cooperative}. 

While the CoVOR-SLAM method provides an innovative perspective on a good trade-off between the estimation accuracy and the computational and communication loads, some other previous studies on the fusion of visual-ranging measurements should be mentioned. 
Wang et al. and Shi et al. used fusion of Ultra-wideband (UWB) ranging between a single robot and an anchor station to mitigate the drift in visual SLAM \cite{wang2017ultra, shi2018visual}.
Xu et al. proposed to fuse visual-inertial and range measurements for multi-agent systems in \cite{xu2020decentralized}. They tested the system using real images and ranges obtained with five drones in an indoor environment. However, they had to first calibrate their system using a pioneer drone to allocate the UWB anchors, and did not exploit the links between the drones. 
Ziegler et al. \cite{ziegler2021distributed} proposed a multi-agent fusion algorithm using Visual-INertial System (VINS) and inter-agent ranges. They tested the system using 49 drones in scenarios synthetically generated using Gazebo \cite{koenig2004design} and RotorS \cite{furrer10robot}. Both odometry and range measurements were synthetically generated using the ground truth with Gaussian noise. 
Queralta et al. additionally used Lidar measurements in \cite{queralta2020vio} to reconstruct the dense scenes, which increased the cost and complexity of the system. 

Compared to other visual-ranging fusion methods, CoVOR-SLAM has the following advantages:
\begin{itemize}
    \item It uses sparse range measurements to accurately estimate the multiple vehicles' poses without scale ambiguity, even when monocular cameras are used (without any other sensors, such as IMU and Lidar). Such additional sensors can be however very easily integrated into the graph optimization based framework. 
    \item In the framework of CoVOR-SLAM, we parameterize the robots' motion as the 7 Degrees of Freedom (DoF) similarity transformation, so the scale error of vision-only positioning can be also considered in the visual-ranging data fusion process (and thus reduced in the fusion). 
    \item It needs much less communication capabilities than inter-agent loop closing, so can be more useful when the communication capacity is limited. In addition, the algorithm can be run on every agent since it does not require large computing power. Thus, it can be useful for decentralized swarm intelligence systems as \cite{zhang2020self}.    
    \item It does not constrain the robots' movement as long as ranges can be measured, which is typically possible even at large distance using e.g. the radio signals between two modules. 
    \item The platforms of the robots can be flexibly chosen, because each swarm element does not need to have similar view points for detecting inter-agent loops in CoVOR-SLAM. 
    \item The algorithm does not require simultaneous connections to multiple anchor stations. Even when one of the rovers is connected to a single anchor station, it can significantly reduce the pose estimation errors of the swarm system. 
\end{itemize}
The system is evaluated using both simulation and experimental data. For the simulations, we generated the ranges including realistic ranging errors modeled using real UWB measurements.

This paper is structured as follows: In \secref{sec:covor_slam_methodology}, we first show the overview of CoVOR-SLAM, and introduce the system model, the measurement model, and the sensor fusion methodology. Then, we show the results of system tests using experimental data in \secref{sec:single_exp}. \secref{sec:system_outdoor_test} presents more results of the system analyses using a larger swarm setup with public datasets. In this section, we show the pose estimation performance and communication requirements of our method (CoVOR-SLAM) compared to the state-of-the-art inter-agent loop closing technique. Finally, conclusions are drawn in \secref{sec:conclusion}.

\section{Cooperative SLAM using Visual Odometry and Ranges}
\label{sec:covor_slam_methodology}

 \begin{figure*}[t!]
   \centering
  \includegraphics[width=0.9\linewidth]{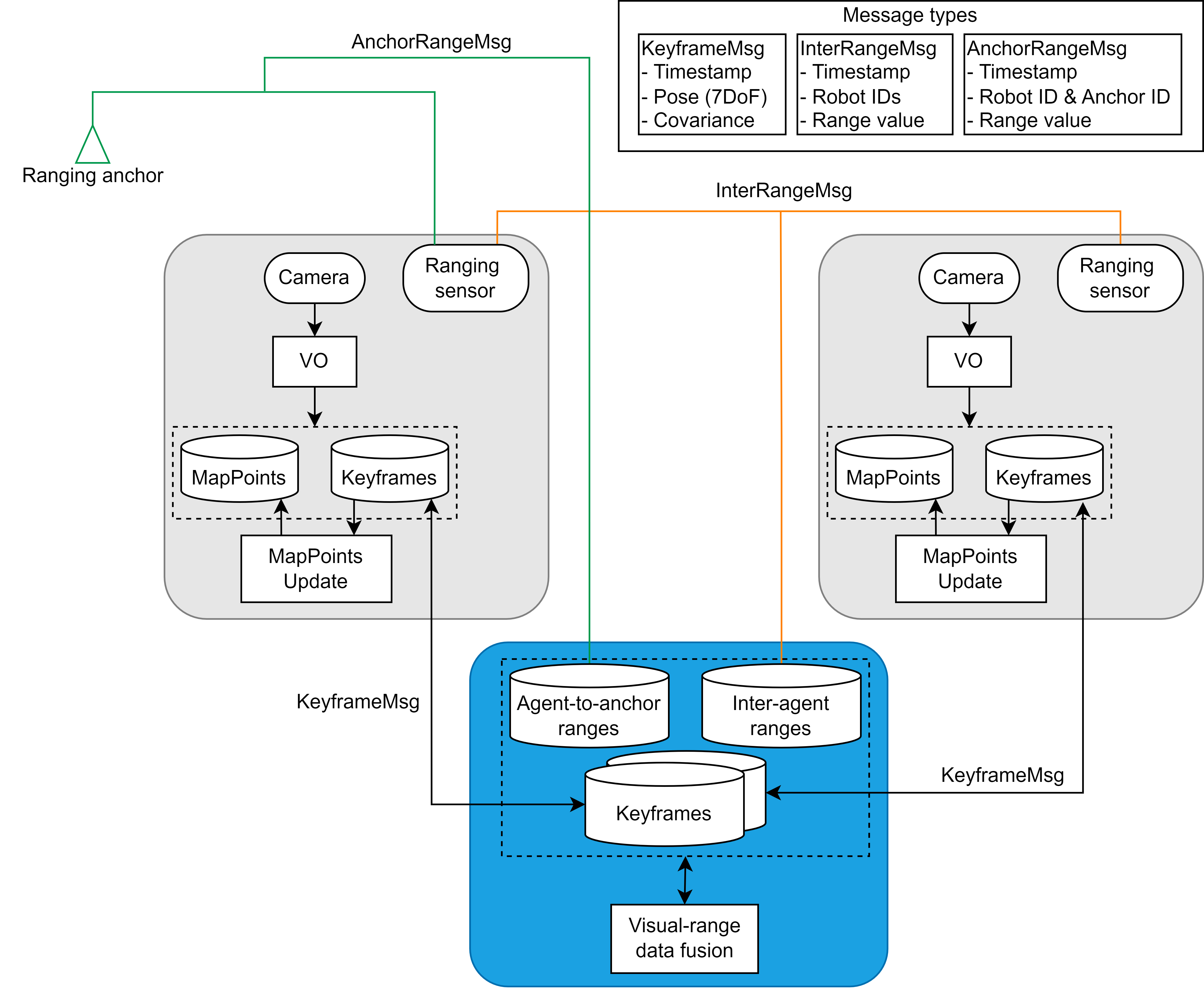}
   \caption{The overview of CoVOR-SLAM}
   \label{fig:multi_overview}
 \end{figure*}

The CoVOR-SLAM framework is shown in \figref{fig:multi_overview}. Each robot estimates its poses as well as map points with respect to its local reference frame using Visual Odometry (VO). The onboard processes are depicted in the gray boxes. 
For the multi-agent data fusion, three types of messages (KeyframeMsg, InterRangeMsg, and AnchorRangeMsg) are exchanged in the data fusion module (blue box). The blue box is drawn separately to explain the system more clearly, but it can be implemented on each agent's onboard processor when the swarm robots are used in a decentralized manner.
The message "KeyframeMsg" contains the robots' poses, which are defined as the 7DoF similarity matrices of the Lie group $\mathrm{Sim(3)}$. A similarity matrix includes the rotation matrix ($\mathbf{R}\in\mathrm{SO(3)}$), translation vector ($\vec{t}{}{}{}\in\mathbb{R}^3$), as well as the scale factor $s$ (scalar value): 
\begin{align*}
    \mat{S}{}{} &= 
    \begin{bmatrix}
        \mat{R}{}{} & \vec{t}{}{}{}\\
        0 & \invs{}{}
    \end{bmatrix} 
    \in \mathrm{Sim(3)}, \\
    \text{\quad where } &\mathbf{R}\in\mathrm{SO(3)}, \vec{t}{}{}{}\in\mathbb{R}^3, s\in\mathbb{R}^1.
\end{align*}
We parameterize the robots' poses as using 7DoF transformations to additionally consider the scale drift of the monocular visual odometry process. 
In addition, we obtain inter-agent ranges using the communication links between the onboard ranging sensors. They are transmitted to the fusion module in the messages "InterRangeMsg" (orange in \figref{fig:multi_overview}). 
Ranges between the robots and the anchor points are also used in the data fusion when agent-to-anchor connections are available ("AnchorRangeMsg", green in \figref{fig:multi_overview}). 
The collected measurements are fused using a graph-based approach to mitigate errors of the robots' pose estimation. 
Furthermore, each robot updates its local mapping using the differences between the keyframe poses before and after the data fusion. 


\subsection{System Model}
\label{subsec:system_model}
\begin{figure}[t!]
\centering
\includegraphics[width=0.9\linewidth]{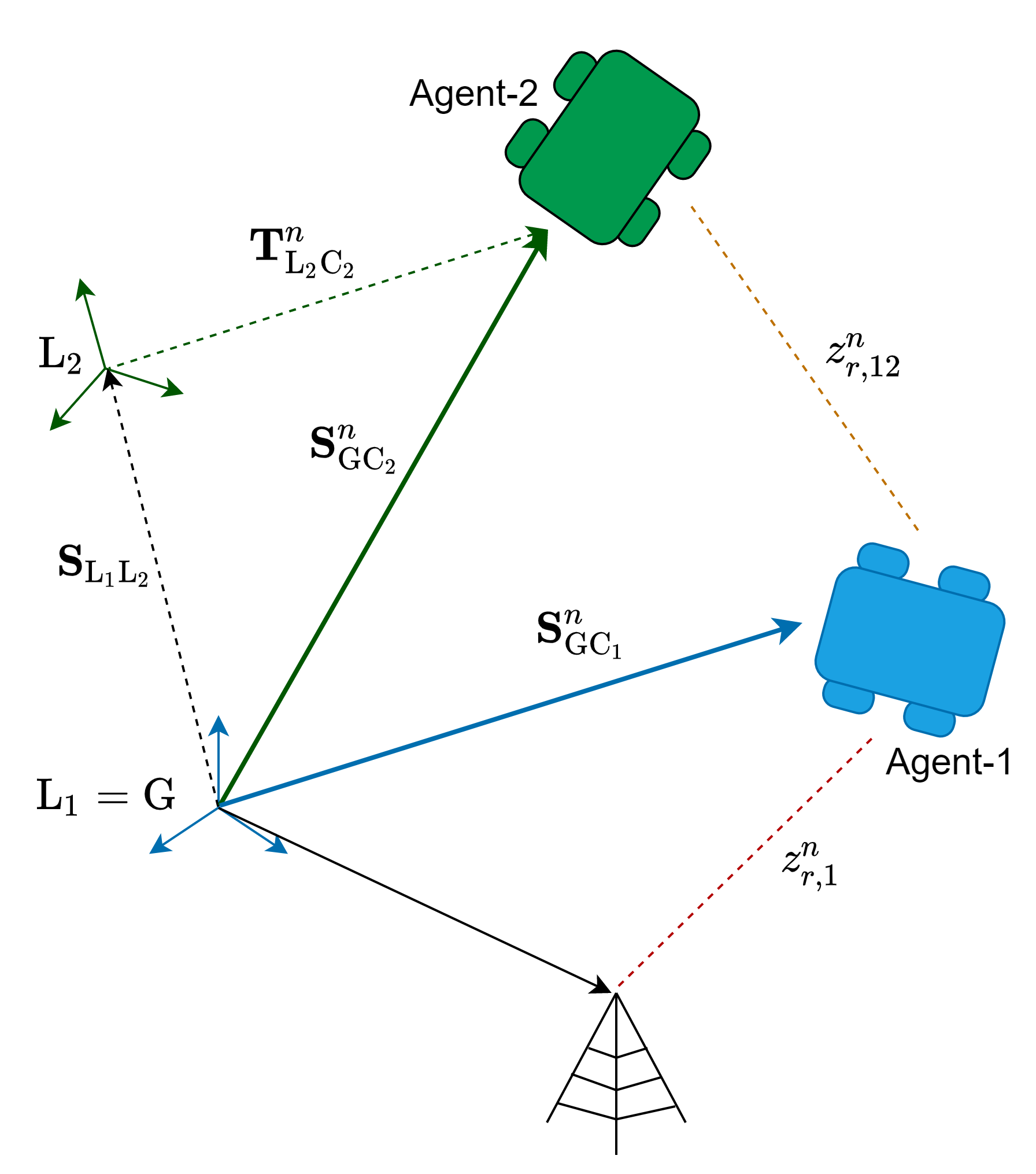}
\caption{The system setup of CoVOR-SLAM with two agents} 
\label{fig:multi_setup}
\end{figure}

The multi-agent system model is shown in \figref{fig:multi_setup}. Two agents are used in the figure to explain the model. 
The navigation frame of agent $k$ is denoted as $\mathrm{L}_k$, which is a fixed frame that the agent uses to denote its own trajectory. Using VO, each robot incrementally estimates the 6DoF poses (3D rotation and translation) of its onboard camera $\mathrm{C}_k$ in its local reference frame. The error-free camera pose at time $n$ in frame $\mathrm{L}_k$ is denoted as: 
\begin{align}
  \mat{T}{L_kC_k}{n} &= 
    \begin{bmatrix}
        \mat{R}{L_kC_k}{n} & \vec{t}{L_kC_k}{n}{}\\
        0 & 1
    \end{bmatrix} 
    \in \mathrm{SE(3)}, \nonumber \\
    \text{\quad where } &\mat{R}{L_kC_k}{n}\in\mathrm{SO(3)}, \vec{t}{L_kC_k}{n}{}\in\mathbb{R}^3.   \label{eq:se3_matrix}
\end{align}
For agents with a monocular camera onboard, the translation vector $\vec{t}{L_kC_k}{n}{}$ can only be estimated with a global scale ambiguity by VO. Consequently, the poses in frame $\mathrm{L}_k$ are also up-to-scale in that case. 

In this exemplary setup, the reference frame of agent-1 is set as the global reference frame (i.e. $\mathrm{L}_1$=G). Agent-1's 6DoF local poses $\mat{T}{L_1C_1}{n}$ with scale ambiguity are thus converted to the 7DoF denotation of the camera states in the global frame $\mat{S}{GC_1}{n}$ by simply adding a scale factor $s_1$. This conversion can be written in matrix form: 
\begin{align}
 \mat{S}{GC_1}{n} = \begin{bmatrix}
 					   \identityMat{3} & \zeroMat{3}{1} \\
 					   \zeroMat{1}{3}  & \invs{1}{}
                    \end{bmatrix}
                    \mat{T}{L_1C_1}{n}  
                  = \mat{S}{GL_1}{} \mat{T}{L_1C_1}{n}.
\label{eqn:transform_L1_G}
\end{align}
When we use a monocular camera to estimate each robot's poses, $s_1$ represents the absolute scale of the monocular camera trajectory estimates. When a stereo camera is used to estimate the poses, we can simply initialize $s_1=1$ because the stereo visual odometry does not have the global scale ambiguity (the absolute scale of the trajectory is estimated using the known baseline length of the stereo camera). Nevertheless, the relative scale estimation error also accumulates over time for stereo cameras.
 
In the global frame G, the poses of Agent-2 $\mat{S}{GC_2}{n}$ (parameterized by 7DoF matrices in $\mathrm{Sim(3)}$) can be transformed from its local poses $\mat{T}{L_2C_2}{n}$ (with scale ambiguity) as: 
\begin{align}
 \mat{S}{GC_2}{n} = \mat{S}{GL_1}{}\mat{S}{L_1L_2}{}\mat{T}{L_2C_2}{n}, 
\end{align}
where $\mat{S}{L_1L_2}{}$ denotes the similarity transformation between the two local reference frames, and $\mat{S}{GL_1}{}$ transforms poses from the frame $\mathrm{L}_1$ to the global frame G as defined in Eqn. (\ref{eqn:transform_L1_G}). $\mat{S}{L_1L_2}{}$ can be estimated during the initialization phase of the mission using a known pattern, such as ArUco markers \cite{garrido2016generation, romero2018speeded} and chessboard. Optionally, the method in \cite{zhu2018journal} can be used to estimate $\mat{S}{L_1L_2}{}$, if both rovers are moving in the same 2D plane during the initialization.

The 7DoF poses of all $K$ agents in the common global frame G are defined as the system state variables to be estimated:
\begin{align} 
\label{eq:ch5_multi_unknowns}
 \unknown = \{ \{ \mat{S}{GC_k}{n_0}, \dots, \mat{S}{GC_k}{n_N} \}, \forall k \in \{ 1,2, \dots, K \} \}.
\end{align}
The timestamps $\{ n_0, \dots , n_N \}$ can be asynchronous for different agents.

\subsection{Measurement Models}
Inter-agent range measurements $\measRange{,12}{n}$ (orange in \figref{fig:multi_setup}) are modeled as the true range between the two onboard ranging sensors (antennas)  $||\vec{p}{12}{n}{}||$ with additive error $\errfnRange{,12}{n}$:
\begin{align}
  \measRange{,12}{n} = || \vec{p}{12}{n}{} || + \errfnRange{,12}{n}.
  \label{eq:inter_agent_range_error}
\end{align}
The offset between the antenna and the camera can be obtained in an offline calibration. Consequently, the position difference in the local reference frame can be compensated. For description clarity, in the following part of the paper, the offsets are assumed to be compensated, so the camera position and the antenna position are denoted with the same symbol.

If available, we also use the ranges between the onboard ranging sensors and anchor points in the data fusion process. For example, ranging between agent-1 and the anchor is available via radio links in \figref{fig:multi_setup} (red dotted line). Similar to the inter-agent links, the anchor-to-agent range measurements are modeled as the true range between the anchor (A) and the onboard ranging-tag module with an additive error: 
\begin{align}
  \measRange{,1}{n} = ||\vec{p}{1A}{n}{}||  + \errfnRange{,1}{n}.
  \label{eq:agent_to_anchor_range_model} 
\end{align}

In addition, we parameterize the odometry measurements of agent $k$ as the estimated 6DoF relative pose at two consecutive timestamps $n_{i}$ and $n_{j}$ using the onboard camera:
\begin{align}
  \measOdo{,k}{ij}
  = & \errMatOdo{,k}{ij} \left( \mat{T}{L_kC_k}{n_j}\invmat{T}{L_kC_k}{n_i} \right) \nonumber \\
  = &  \underbrace{ \errMatOdo{,k}{ij} }_{\substack{\text{VO error in}\\ \text{matrix form}}} 
  \underbrace{ \left( \mat{S}{GC_k}{n_j}\invmat{S}{GC_k}{n_i} \right) }_{\substack{\text{Relative motion between}\\ \text{time $n_i$ and $n_j$}}} 
\label{eq:odo_meas_model}
\end{align}
The global frame poses and the visual odometry errors are parameterized in 7DoF matrix form, i.e., $\mat{S}{GC_k}{n_i}, \mat{S}{GC_k}{n_j}, \errMatOdo{k}{ij}\in\mathrm{Sim(3)}$, so that the scale drift error is taken into account.


\subsection{Multi-Agent Visual-Range Data Fusion}
\label{subsec:multi_agent_data_fusion}
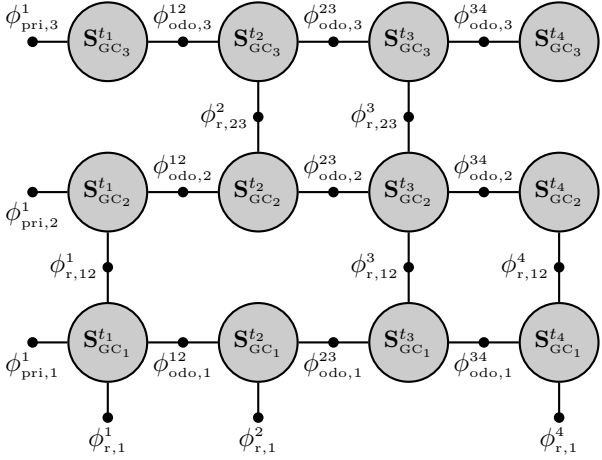
\begin{figure}[t!]
    \centering
    \begin{tikzpicture}[auto, node distance=3cm, every loop/.style={},
                    thick,main node/.style={circle,draw,font=\sffamily\Large\bfseries}]
 
  \usetikzlibrary{decorations.markings}
  
  \def\hnode{2}
  \def\vnode{2}
  \def\dagent{2}
  
  \def\posescale{0.9}

  \node (c1_1) at (\hnode,0) [circle, fill=black!20, scale=\posescale, draw] {$\mat{S}{GC_1}{t_1}$};
  \node (c1_2) at (\hnode*2,0) [circle, fill=black!20, scale=\posescale, draw] {$\mat{S}{GC_1}{t_2}$};
  \node (c1_3) at (\hnode*3,0) [circle, fill=black!20, scale=\posescale,draw] {$\mat{S}{GC_1}{t_3}$};
  \node (c1_4) at (\hnode*4,0) [circle, fill=black!20, scale=\posescale,draw] {$\mat{S}{GC_1}{t_4}$};
  
  \node (c2_1) at (\hnode,\vnode) [circle, fill=black!20, scale=\posescale, draw] {$\mat{S}{GC_2}{t_1}$};
  \node (c2_2) at (\hnode*2,\vnode) [circle, fill=black!20, scale=\posescale, draw] {$\mat{S}{GC_2}{t_2}$};
  \node (c2_3) at (\hnode*3,\vnode) [circle, fill=black!20, scale=\posescale,draw] {$\mat{S}{GC_2}{t_3}$};
  \node (c2_4) at (\hnode*4,\vnode) [circle, fill=black!20, scale=\posescale,draw] {$\mat{S}{GC_2}{t_4}$};
  
   \node (c3_1) at (\hnode,\vnode*2) [circle, fill=black!20, scale=\posescale, draw] {$\mat{S}{GC_3}{t_1}$};
   \node (c3_2) at (\hnode*2,\vnode*2) [circle, fill=black!20, scale=\posescale, draw] {$\mat{S}{GC_3}{t_2}$};
   \node (c3_3) at (\hnode*3,\vnode*2) [circle, fill=black!20, scale=\posescale,draw] {$\mat{S}{GC_3}{t_3}$};
   \node (c3_4) at (\hnode*4,\vnode*2) [circle, fill=black!20, scale=\posescale,draw] {$\mat{S}{GC_3}{t_4}$};
%
  
  
  \unaryfactor{{pri,1}}{1}{c1_1}{(\hnode/2,0)}{below};
 
  \binaryfactor{{odo,1}}{12}{c1_1}{c1_2}{(\hnode+\hnode/2,0)}{below};
  \binaryfactor{{odo,1}}{23}{c1_2}{c1_3}{(\hnode*2+\hnode/2,0)}{below};
  \binaryfactor{{odo,1}}{34}{c1_3}{c1_4}{(\hnode*3+\hnode/2,0)}{below};
  
  \unaryfactor{{pri,2}}{1}{c2_1}{(\hnode/2,\vnode)}{below};

  \binaryfactor{{odo,2}}{12}{c2_1}{c2_2}{(\hnode+\hnode/2,\vnode)}{above};
  \binaryfactor{{odo,2}}{23}{c2_2}{c2_3}{(\hnode*2+\hnode/2,\vnode)}{above};
  \binaryfactor{{odo,2}}{34}{c2_3}{c2_4}{(\hnode*3+\hnode/2,\vnode)}{above};
  
   \unaryfactor{{pri,3}}{1}{c3_1}{(\hnode/2,\vnode*2)}{above};
 
   \binaryfactor{{odo,3}}{12}{c3_1}{c3_2}{(\hnode+\hnode/2,\vnode*2)}{above};
   \binaryfactor{{odo,3}}{23}{c3_2}{c3_3}{(\hnode*2+\hnode/2,\vnode*2)}{above};
   \binaryfactor{{odo,3}}{34}{c3_3}{c3_4}{(\hnode*3+\hnode/2,\vnode*2)}{above};
   
%
%
%
  
  \binaryfactor{r,12}{1}{c1_1}{c2_1}{(\hnode,\vnode/2)}{left};
  \binaryfactor{r,12}{3}{c1_3}{c2_3}{(\hnode*3,\vnode/2)}{left};
  \binaryfactor{r,12}{4}{c1_4}{c2_4}{(\hnode*4,\vnode/2)}{left};
  
   \binaryfactor{r,23}{2}{c2_2}{c3_2}{(\hnode*2,\vnode*1.5)}{left};
   \binaryfactor{r,23}{3}{c2_3}{c3_3}{(\hnode*3,\vnode*1.5)}{left};
   

%

  \unaryfactor{r,1}{1}{c1_1}{(\hnode,-\hnode/2.0)}{below};
  \unaryfactor{r,1}{2}{c1_2}{(\hnode*2.0,-\hnode/2.0)}{below};
  \unaryfactor{r,1}{4}{c1_4}{(\hnode*4.0,-\hnode/2.0)}{below};

\end{tikzpicture}
    \caption{Factor graph created for fusing visual odometry and range measurements of multiple agents (created with the three-agent setup)}.
    \label{fig:multi_factorgraph}
\end{figure}

After collecting all the measurements, we create a factor graph as shown in \figref{fig:multi_factorgraph}. The figure shows an example of the factor graph created employing a three-agent setup (with each agent's poses in a row). 
In this graph, the state variables (the camera poses $\mat{S}{GC_1}{n}$) are depicted as the circular nodes, and the measurement factors $\factornode{}{}$ are added between or at the nodes.  

The factors $\factornode{}{}$ are the likelihood of the states, given each measurement. For example, the inter-agent range factor $\factornodeRange{,k\kprime}{i}$ added between agent-$k$ and agent-$\kprime$ at $n_i$ is the likelihood function of the 7DoF poses of two agents at $n_i$, given the distribution of inter-agent range measurement $n_i$, i.e. $ \factornodeRange{,k\kprime}{i}=l(\mat{S}{GC_k}{n_i},\mat{S}{GC_\kprime}{n_i}|\measRange{,k\kprime}{n_i})$. 

In this graph, we assume that only agent-1 receives the signals from the anchor at time $n_1$, $n_2$, and $n_4$, and the others are not connected to the anchor point, so the anchor range factors $\factornodeRange{,k}{i}$ (index $i$ for time $n_i$) are only added to agent-1's nodes ($k=1$). 
In addition, the odometry factors $\factornodeOdo{,k}{ij}$ are added between the two consecutive poses of each agent, and the prior factor $\factornodePri{,k}{1}$ is also added at each agent's pose at the starting time. 

Finally, a Maximum A Posteriori (MAP) problem is formulated, using all the factors in the graph: 
\begin{align}
  \unknownOpt = \argmax{\unknown} \prod_{k,\kprime}\prod_{i} \factornodeRange{,k\kprime}{i}
                                  \prod_{k}\prod_{i} \factornodeRange{,k}{i}
                                  \prod_{k}\prod_{i,j} \factornodeOdo{,k}{ij}
                                  \prod_{k}\factornodePri{,k}{1} .
\end{align}
$\unknown$ contains the poses to be estimated as defined in Eqn. (\ref{eq:ch5_multi_unknowns}), parameterized by matrix elements in Lie group $\mathrm{Sim(3)}$.

If all the measurement noise is Gaussian, the MAP problem is equivalent to the Least-Squares (LS) problem that minimizes the sum of the squared measurement errors as
\begin{align}
\unknownOpt = &\argmin{\unknown}
\underbrace{\sum_{k,\kprime}\sum_{i} \transpose{\errfnRange{,k\kprime}{i}} \mathbf{W}_{\text{r},k k'}^i \errfnRange{,k\kprime}{i}}_\text{Inter-agent range errors}   \nonumber \\
&+\underbrace{\sum_{k}\sum_{i}\transpose{\errfnRange{,k}{i}} \mathbf{W}_{\text{r},k}^i \errfnRange{,k}{i}}_\text{Agent-to-anchor range errors}  + \underbrace{\sum_{k}\transpose{\errfnPri{,k}} \mathbf{W}_{\text{pri},k} \errfnPri{,k}}_\text{A prior error} \nonumber \\
&+ \underbrace{\sum_{k}\sum_{i,j}\transpose{\errfnOdo{,k}{ij}}  \mathbf{W}_{\text{odo},k}^{ij} \errfnOdo{,k}{ij}}_\text{Odometry errors}. 
\label{eqn:SSE_measurements}
\end{align}
$\mathbf{W}_{*,k}^i$ denotes the inverse of covariance matrix of the corrsponding measurements of agent $k$ at time $n_i$ (for simplicity we use $k$ and $i$ to represent all the different agent indices and time indices in general). The odometry errors $\errfnOdo{}{ij}$ in Eqn. (\ref{eqn:SSE_measurements}) is the $7\times1$ coordinates converted from the  $4\times4$ error matrices $ \errMatOdo{}{ij}$ (see \eqref{eq:odo_meas_model}), using the logarithmic relationship between the twist coordinates and Lie group matrices as explained in \cite{hall2015lie}:
\begin{align}
  \errfnOdo{}{ij} 
  =\ln\left( \errMatOdo{}{ij} \right)  
  = \ln\left( \measOdo{}{ij}\left( \mat{S}{GC}{n_i}\invmat{S}{GC}{n_j} \right)  \right). \label{eq:single_odo_errVec}
\end{align}

\begin{figure}
\centering
\captionsetup[subfloat]{labelfont=normalsize,textfont=normalsize}
\subfloat[A prior]{\tikzset{every picture/.style={line width=0.75pt}} 

\begin{tikzpicture}[x=0.75pt,y=0.75pt,yscale=-1,xscale=1]

\draw  [draw opacity=0] (164.22,288.99) -- (247.22,288.99) -- (247.22,371.99) -- (164.22,371.99) -- cycle ; \draw  [color={rgb, 255:red, 74; green, 74; blue, 74 }  ,draw opacity=1 ] (171.14,288.99) -- (171.14,371.99)(178.05,288.99) -- (178.05,371.99)(184.97,288.99) -- (184.97,371.99)(191.89,288.99) -- (191.89,371.99)(198.8,288.99) -- (198.8,371.99)(205.72,288.99) -- (205.72,371.99)(212.64,288.99) -- (212.64,371.99)(219.55,288.99) -- (219.55,371.99)(226.47,288.99) -- (226.47,371.99)(233.39,288.99) -- (233.39,371.99)(240.3,288.99) -- (240.3,371.99) ; \draw  [color={rgb, 255:red, 74; green, 74; blue, 74 }  ,draw opacity=1 ] (164.22,295.91) -- (247.22,295.91)(164.22,302.83) -- (247.22,302.83)(164.22,309.74) -- (247.22,309.74)(164.22,316.66) -- (247.22,316.66)(164.22,323.58) -- (247.22,323.58)(164.22,330.49) -- (247.22,330.49)(164.22,337.41) -- (247.22,337.41)(164.22,344.33) -- (247.22,344.33)(164.22,351.24) -- (247.22,351.24)(164.22,358.16) -- (247.22,358.16)(164.22,365.08) -- (247.22,365.08) ; \draw  [color={rgb, 255:red, 74; green, 74; blue, 74 }  ,draw opacity=1 ] (164.22,288.99) -- (247.22,288.99) -- (247.22,371.99) -- (164.22,371.99) -- cycle ;
\draw [color={rgb, 255:red, 74; green, 74; blue, 74 }  ,draw opacity=1 ][line width=1.5]    (164.22,316.47) -- (246.64,315.91) ;
\draw [color={rgb, 255:red, 74; green, 74; blue, 74 }  ,draw opacity=1 ][line width=1.5]    (164.22,343.61) -- (246.64,343.61) ;
\draw  [draw opacity=0][fill={rgb, 255:red, 155; green, 155; blue, 155 }  ,fill opacity=1 ] (164.22,289.33) -- (171.09,289.33) -- (171.09,296.06) -- (164.22,296.06) -- cycle ;
\draw  [draw opacity=0][fill={rgb, 255:red, 155; green, 155; blue, 155 }  ,fill opacity=1 ] (191.7,317.24) -- (198.56,317.24) -- (198.56,323.97) -- (191.7,323.97) -- cycle ;
\draw  [draw opacity=0][fill={rgb, 255:red, 155; green, 155; blue, 155 }  ,fill opacity=1 ] (219.17,344.16) -- (226.04,344.16) -- (226.04,350.89) -- (219.17,350.89) -- cycle ;
\draw [color={rgb, 255:red, 74; green, 74; blue, 74 }  ,draw opacity=1 ][line width=1.5]    (191.89,288.99) -- (191.89,371.99) ;
\draw [color={rgb, 255:red, 74; green, 74; blue, 74 }  ,draw opacity=1 ][line width=1.5]    (219.17,288.99) -- (219.55,371.99) ;

\draw (159.06,355.5) node  [font=\tiny,rotate=-270] [align=left] {Agent-3};
\draw (159.06,327.47) node  [font=\tiny,rotate=-270] [align=left] {Agent-2};
\draw (159.06,299.99) node  [font=\tiny,rotate=-270] [align=left] {Agent-1};
\draw (235.22,284.42) node  [font=\tiny] [align=left] {Agent-3};
\draw (207.18,284.42) node  [font=\tiny] [align=left] {Agent-2};
\draw (178.56,284.42) node  [font=\tiny] [align=left] {Agent-1};

\end{tikzpicture}\label{subfig:Hp_structure}} ~
\subfloat[Odometry]{\tikzset{every picture/.style={line width=0.75pt}} 

\begin{tikzpicture}[x=0.75pt,y=0.75pt,yscale=-1,xscale=1]

\draw  [draw opacity=0] (333.54,288.12) -- (416.54,288.12) -- (416.54,371.12) -- (333.54,371.12) -- cycle ; \draw  [color={rgb, 255:red, 74; green, 74; blue, 74 }  ,draw opacity=1 ] (340.46,288.12) -- (340.46,371.12)(347.38,288.12) -- (347.38,371.12)(354.29,288.12) -- (354.29,371.12)(361.21,288.12) -- (361.21,371.12)(368.13,288.12) -- (368.13,371.12)(375.04,288.12) -- (375.04,371.12)(381.96,288.12) -- (381.96,371.12)(388.88,288.12) -- (388.88,371.12)(395.79,288.12) -- (395.79,371.12)(402.71,288.12) -- (402.71,371.12)(409.63,288.12) -- (409.63,371.12) ; \draw  [color={rgb, 255:red, 74; green, 74; blue, 74 }  ,draw opacity=1 ] (333.54,295.03) -- (416.54,295.03)(333.54,301.95) -- (416.54,301.95)(333.54,308.87) -- (416.54,308.87)(333.54,315.78) -- (416.54,315.78)(333.54,322.7) -- (416.54,322.7)(333.54,329.62) -- (416.54,329.62)(333.54,336.53) -- (416.54,336.53)(333.54,343.45) -- (416.54,343.45)(333.54,350.37) -- (416.54,350.37)(333.54,357.28) -- (416.54,357.28)(333.54,364.2) -- (416.54,364.2) ; \draw  [color={rgb, 255:red, 74; green, 74; blue, 74 }  ,draw opacity=1 ] (333.54,288.12) -- (416.54,288.12) -- (416.54,371.12) -- (333.54,371.12) -- cycle ;
\draw  [draw opacity=0][fill={rgb, 255:red, 155; green, 155; blue, 155 }  ,fill opacity=1 ] (333.54,288.12) -- (361.02,288.12) -- (361.02,315.03) -- (333.54,315.03) -- cycle ;

\draw  [draw opacity=0][fill={rgb, 255:red, 155; green, 155; blue, 155 }  ,fill opacity=1 ] (361.02,315.03) -- (388.88,315.03) -- (388.88,343.45) -- (361.02,343.45) -- cycle ;
\draw  [draw opacity=0][fill={rgb, 255:red, 155; green, 155; blue, 155 }  ,fill opacity=1 ] (388.49,341.94) -- (416.54,341.94) -- (416.54,371.12) -- (388.49,371.12) -- cycle ;
\draw [color={rgb, 255:red, 74; green, 74; blue, 74 }  ,draw opacity=1 ][line width=1.5]    (361.21,288.12) -- (361.21,371.12) ;
\draw [color={rgb, 255:red, 74; green, 74; blue, 74 }  ,draw opacity=1 ][line width=1.5]    (333.54,315.59) -- (415.96,315.03) ;
\draw [color={rgb, 255:red, 74; green, 74; blue, 74 }  ,draw opacity=1 ][line width=1.5]    (333.54,342.94) -- (415.96,342.94) ;
\draw [color={rgb, 255:red, 74; green, 74; blue, 74 }  ,draw opacity=1 ][line width=1.5]    (388.88,288.12) -- (388.88,371.12) ;

\draw (328.38,299.12) node  [font=\tiny,rotate=-270] [align=left] {Agent-1};
\draw (328.38,326.59) node  [font=\tiny,rotate=-270] [align=left] {Agent-2};
\draw (328.38,354.62) node  [font=\tiny,rotate=-270] [align=left] {Agent-3};
\draw (404.55,283.54) node  [font=\tiny] [align=left] {Agent-3};
\draw (376.5,283.54) node  [font=\tiny] [align=left] {Agent-2};
\draw (347.88,283.54) node  [font=\tiny] [align=left] {Agent-1};

\end{tikzpicture}\label{subfig:Hodo_structure}} \\
\subfloat[Agent-to-anchor range]{\tikzset{every picture/.style={line width=0.75pt}} 

\begin{tikzpicture}[x=0.75pt,y=0.75pt,yscale=-1,xscale=1]

\draw  [draw opacity=0] (486.73,290.24) -- (569.73,290.24) -- (569.73,373.24) -- (486.73,373.24) -- cycle ; \draw  [color={rgb, 255:red, 74; green, 74; blue, 74 }  ,draw opacity=1 ] (493.64,290.24) -- (493.64,373.24)(500.56,290.24) -- (500.56,373.24)(507.48,290.24) -- (507.48,373.24)(514.39,290.24) -- (514.39,373.24)(521.31,290.24) -- (521.31,373.24)(528.23,290.24) -- (528.23,373.24)(535.14,290.24) -- (535.14,373.24)(542.06,290.24) -- (542.06,373.24)(548.98,290.24) -- (548.98,373.24)(555.89,290.24) -- (555.89,373.24)(562.81,290.24) -- (562.81,373.24) ; \draw  [color={rgb, 255:red, 74; green, 74; blue, 74 }  ,draw opacity=1 ] (486.73,297.15) -- (569.73,297.15)(486.73,304.07) -- (569.73,304.07)(486.73,310.99) -- (569.73,310.99)(486.73,317.9) -- (569.73,317.9)(486.73,324.82) -- (569.73,324.82)(486.73,331.74) -- (569.73,331.74)(486.73,338.65) -- (569.73,338.65)(486.73,345.57) -- (569.73,345.57)(486.73,352.49) -- (569.73,352.49)(486.73,359.4) -- (569.73,359.4)(486.73,366.32) -- (569.73,366.32) ; \draw  [color={rgb, 255:red, 74; green, 74; blue, 74 }  ,draw opacity=1 ] (486.73,290.24) -- (569.73,290.24) -- (569.73,373.24) -- (486.73,373.24) -- cycle ;
\draw [color={rgb, 255:red, 74; green, 74; blue, 74 }  ,draw opacity=1 ][line width=1.5]    (486.73,317.71) -- (569.15,317.15) ;
\draw [color={rgb, 255:red, 74; green, 74; blue, 74 }  ,draw opacity=1 ][line width=1.5]    (486.73,345.06) -- (569.15,345.06) ;
\draw [color={rgb, 255:red, 74; green, 74; blue, 74 }  ,draw opacity=1 ][line width=1.5]    (541.67,290.24) -- (542.06,373.24) ;

\draw  [draw opacity=0][fill={rgb, 255:red, 155; green, 155; blue, 155 }  ,fill opacity=1 ] (486.73,290.24) -- (493.59,290.24) -- (493.59,296.97) -- (486.73,296.97) -- cycle ;
\draw  [draw opacity=0][fill={rgb, 255:red, 155; green, 155; blue, 155 }  ,fill opacity=1 ] (493.59,296.97) -- (500.46,296.97) -- (500.46,303.69) -- (493.59,303.69) -- cycle ;
\draw  [draw opacity=0][fill={rgb, 255:red, 155; green, 155; blue, 155 }  ,fill opacity=1 ] (507.33,310.42) -- (514.2,310.42) -- (514.2,317.15) -- (507.33,317.15) -- cycle ;
\draw [color={rgb, 255:red, 74; green, 74; blue, 74 }  ,draw opacity=1 ][line width=1.5]    (514.77,290.24) -- (514.39,373.24) ;

\draw (481.56,301.24) node  [font=\tiny,rotate=-270] [align=left] {Agent-1};
\draw (481.56,328.71) node  [font=\tiny,rotate=-270] [align=left] {Agent-2};
\draw (481.56,356.74) node  [font=\tiny,rotate=-270] [align=left] {Agent-3};
\draw (557.73,285.66) node  [font=\tiny] [align=left] {Agent-3};
\draw (529.68,285.66) node  [font=\tiny] [align=left] {Agent-2};
\draw (501.07,285.66) node  [font=\tiny] [align=left] {Agent-1};

\end{tikzpicture}\label{subfig:Hanchor_structure}} ~
\subfloat[Inter-agent range]{\tikzset{every picture/.style={line width=0.75pt}} 

\begin{tikzpicture}[x=0.75pt,y=0.75pt,yscale=-1,xscale=1]

\draw  [draw opacity=0] (271.22,339.99) -- (354.22,339.99) -- (354.22,422.99) -- (271.22,422.99) -- cycle ; \draw  [color={rgb, 255:red, 74; green, 74; blue, 74 }  ,draw opacity=1 ] (278.14,339.99) -- (278.14,422.99)(285.05,339.99) -- (285.05,422.99)(291.97,339.99) -- (291.97,422.99)(298.89,339.99) -- (298.89,422.99)(305.8,339.99) -- (305.8,422.99)(312.72,339.99) -- (312.72,422.99)(319.64,339.99) -- (319.64,422.99)(326.55,339.99) -- (326.55,422.99)(333.47,339.99) -- (333.47,422.99)(340.39,339.99) -- (340.39,422.99)(347.3,339.99) -- (347.3,422.99) ; \draw  [color={rgb, 255:red, 74; green, 74; blue, 74 }  ,draw opacity=1 ] (271.22,346.91) -- (354.22,346.91)(271.22,353.83) -- (354.22,353.83)(271.22,360.74) -- (354.22,360.74)(271.22,367.66) -- (354.22,367.66)(271.22,374.58) -- (354.22,374.58)(271.22,381.49) -- (354.22,381.49)(271.22,388.41) -- (354.22,388.41)(271.22,395.33) -- (354.22,395.33)(271.22,402.24) -- (354.22,402.24)(271.22,409.16) -- (354.22,409.16)(271.22,416.08) -- (354.22,416.08) ; \draw  [color={rgb, 255:red, 74; green, 74; blue, 74 }  ,draw opacity=1 ] (271.22,339.99) -- (354.22,339.99) -- (354.22,422.99) -- (271.22,422.99) -- cycle ;
\draw [color={rgb, 255:red, 74; green, 74; blue, 74 }  ,draw opacity=1 ][line width=1.5]    (271.22,367.47) -- (353.64,366.91) ;
\draw [color={rgb, 255:red, 74; green, 74; blue, 74 }  ,draw opacity=1 ][line width=1.5]    (271.22,394.61) -- (353.64,394.61) ;
\draw [color={rgb, 255:red, 74; green, 74; blue, 74 }  ,draw opacity=1 ][line width=1.5]    (299.27,339.99) -- (298.7,420.73) ;
\draw [color={rgb, 255:red, 74; green, 74; blue, 74 }  ,draw opacity=1 ][line width=1.5]    (326.17,339.99) -- (326.17,420.73) ;
\draw  [draw opacity=0][fill={rgb, 255:red, 155; green, 155; blue, 155 }  ,fill opacity=1 ] (271.22,340.33) -- (278.09,340.33) -- (278.09,347.06) -- (271.22,347.06) -- cycle ;
\draw  [draw opacity=0][fill={rgb, 255:red, 155; green, 155; blue, 155 }  ,fill opacity=1 ] (298.7,367.24) -- (305.56,367.24) -- (305.56,373.97) -- (298.7,373.97) -- cycle ;
\draw  [draw opacity=0][fill={rgb, 255:red, 155; green, 155; blue, 155 }  ,fill opacity=1 ] (326.17,394.16) -- (333.04,394.16) -- (333.04,400.89) -- (326.17,400.89) -- cycle ;

\draw (266.06,406.5) node  [font=\tiny,rotate=-270] [align=left] {Agent-3};
\draw (266.06,378.47) node  [font=\tiny,rotate=-270] [align=left] {Agent-2};
\draw (266.06,350.99) node  [font=\tiny,rotate=-270] [align=left] {Agent-1};
\draw (342.22,335.42) node  [font=\tiny] [align=left] {Agent-3};
\draw (314.18,335.42) node  [font=\tiny] [align=left] {Agent-2};
\draw (285.56,335.42) node  [font=\tiny] [align=left] {Agent-1};

\end{tikzpicture}\label{subfig:Hinter_structure}} ~
\subfloat[Combination]{\tikzset{every picture/.style={line width=0.75pt}} 

\begin{tikzpicture}[x=0.75pt,y=0.75pt,yscale=-1,xscale=1]

\draw  [draw opacity=0] (471.7,442.04) -- (554.7,442.04) -- (554.7,525.04) -- (471.7,525.04) -- cycle ; \draw  [color={rgb, 255:red, 74; green, 74; blue, 74 }  ,draw opacity=1 ] (478.62,442.04) -- (478.62,525.04)(485.53,442.04) -- (485.53,525.04)(492.45,442.04) -- (492.45,525.04)(499.37,442.04) -- (499.37,525.04)(506.28,442.04) -- (506.28,525.04)(513.2,442.04) -- (513.2,525.04)(520.12,442.04) -- (520.12,525.04)(527.03,442.04) -- (527.03,525.04)(533.95,442.04) -- (533.95,525.04)(540.87,442.04) -- (540.87,525.04)(547.78,442.04) -- (547.78,525.04) ; \draw  [color={rgb, 255:red, 74; green, 74; blue, 74 }  ,draw opacity=1 ] (471.7,448.96) -- (554.7,448.96)(471.7,455.87) -- (554.7,455.87)(471.7,462.79) -- (554.7,462.79)(471.7,469.71) -- (554.7,469.71)(471.7,476.62) -- (554.7,476.62)(471.7,483.54) -- (554.7,483.54)(471.7,490.46) -- (554.7,490.46)(471.7,497.37) -- (554.7,497.37)(471.7,504.29) -- (554.7,504.29)(471.7,511.21) -- (554.7,511.21)(471.7,518.12) -- (554.7,518.12) ; \draw  [color={rgb, 255:red, 74; green, 74; blue, 74 }  ,draw opacity=1 ] (471.7,442.04) -- (554.7,442.04) -- (554.7,525.04) -- (471.7,525.04) -- cycle ;
\draw [color={rgb, 255:red, 74; green, 74; blue, 74 }  ,draw opacity=1 ][line width=1.5]    (471.7,469.51) -- (554.12,468.95) ;
\draw [color={rgb, 255:red, 74; green, 74; blue, 74 }  ,draw opacity=1 ][line width=1.5]    (471.7,495.87) -- (554.12,495.87) ;
\draw [color={rgb, 255:red, 74; green, 74; blue, 74 }  ,draw opacity=1 ][line width=1.5]    (499.75,442.04) -- (499.37,525.04) ;
\draw [color={rgb, 255:red, 74; green, 74; blue, 74 }  ,draw opacity=1 ][line width=1.5]    (526.65,442.04) -- (527.03,525.04) ;

\draw  [draw opacity=0][fill={rgb, 255:red, 74; green, 74; blue, 74 }  ,fill opacity=1 ] (471.7,442.04) -- (478.57,442.04) -- (478.57,448.77) -- (471.7,448.77) -- cycle ;
\draw  [draw opacity=0][fill={rgb, 255:red, 74; green, 74; blue, 74 }  ,fill opacity=1 ] (499.17,468.95) -- (506.04,468.95) -- (506.04,475.68) -- (499.17,475.68) -- cycle ;
\draw  [draw opacity=0][fill={rgb, 255:red, 74; green, 74; blue, 74 }  ,fill opacity=1 ] (471.7,469.51) -- (478.57,469.51) -- (478.57,476.24) -- (471.7,476.24) -- cycle ;
\draw  [draw opacity=0][fill={rgb, 255:red, 74; green, 74; blue, 74 }  ,fill opacity=1 ] (499.17,442.04) -- (506.04,442.04) -- (506.04,448.77) -- (499.17,448.77) -- cycle ;
\draw  [draw opacity=0][fill={rgb, 255:red, 74; green, 74; blue, 74 }  ,fill opacity=1 ] (478.57,448.77) -- (485.44,448.77) -- (485.44,455.5) -- (478.57,455.5) -- cycle ;
\draw  [draw opacity=0][fill={rgb, 255:red, 74; green, 74; blue, 74 }  ,fill opacity=1 ] (506.04,475.68) -- (512.91,475.68) -- (512.91,482.41) -- (506.04,482.41) -- cycle ;
\draw  [draw opacity=0][fill={rgb, 255:red, 74; green, 74; blue, 74 }  ,fill opacity=1 ] (478.57,476.68) -- (485.44,476.68) -- (485.44,483.41) -- (478.57,483.41) -- cycle ;
\draw  [draw opacity=0][fill={rgb, 255:red, 74; green, 74; blue, 74 }  ,fill opacity=1 ] (506.04,448.77) -- (512.91,448.77) -- (512.91,455.5) -- (506.04,455.5) -- cycle ;
\draw  [draw opacity=0][fill={rgb, 255:red, 74; green, 74; blue, 74 }  ,fill opacity=1 ] (492.31,462.23) -- (499.17,462.23) -- (499.17,468.95) -- (492.31,468.95) -- cycle ;
\draw  [draw opacity=0][fill={rgb, 255:red, 74; green, 74; blue, 74 }  ,fill opacity=1 ] (519.78,489.14) -- (526.65,489.14) -- (526.65,495.87) -- (519.78,495.87) -- cycle ;
\draw  [draw opacity=0][fill={rgb, 255:red, 74; green, 74; blue, 74 }  ,fill opacity=1 ] (492.31,489.14) -- (499.17,489.14) -- (499.17,495.87) -- (492.31,495.87) -- cycle ;
\draw  [draw opacity=0][fill={rgb, 255:red, 74; green, 74; blue, 74 }  ,fill opacity=1 ] (519.78,462.23) -- (526.65,462.23) -- (526.65,468.95) -- (519.78,468.95) -- cycle ;
\draw  [draw opacity=0][fill={rgb, 255:red, 74; green, 74; blue, 74 }  ,fill opacity=1 ] (533.52,502.6) -- (540.38,502.6) -- (540.38,509.32) -- (533.52,509.32) -- cycle ;
\draw  [draw opacity=0][fill={rgb, 255:red, 74; green, 74; blue, 74 }  ,fill opacity=1 ] (540.38,509.32) -- (547.25,509.32) -- (547.25,516.05) -- (540.38,516.05) -- cycle ;
\draw  [draw opacity=0][fill={rgb, 255:red, 74; green, 74; blue, 74 }  ,fill opacity=1 ] (533.52,476.68) -- (540.38,476.68) -- (540.38,483.41) -- (533.52,483.41) -- cycle ;
\draw  [draw opacity=0][fill={rgb, 255:red, 74; green, 74; blue, 74 }  ,fill opacity=1 ] (540.38,483.41) -- (547.25,483.41) -- (547.25,490.14) -- (540.38,490.14) -- cycle ;
\draw  [draw opacity=0][fill={rgb, 255:red, 74; green, 74; blue, 74 }  ,fill opacity=1 ] (506.04,504.6) -- (512.91,504.6) -- (512.91,511.32) -- (506.04,511.32) -- cycle ;
\draw  [draw opacity=0][fill={rgb, 255:red, 74; green, 74; blue, 74 }  ,fill opacity=1 ] (512.91,511.32) -- (519.78,511.32) -- (519.78,518.05) -- (512.91,518.05) -- cycle ;
\draw  [draw opacity=0][fill={rgb, 255:red, 74; green, 74; blue, 74 }  ,fill opacity=1 ] (471.7,442.04) -- (499.17,442.04) -- (499.17,468.95) -- (471.7,468.95) -- cycle ;
\draw  [draw opacity=0][fill={rgb, 255:red, 74; green, 74; blue, 74 }  ,fill opacity=1 ] (499.17,468.95) -- (526.65,468.95) -- (526.65,495.87) -- (499.17,495.87) -- cycle ;
\draw  [draw opacity=0][fill={rgb, 255:red, 74; green, 74; blue, 74 }  ,fill opacity=1 ] (526.65,497.37) -- (554.7,497.37) -- (554.7,525.04) -- (526.65,525.04) -- cycle ;

\draw (466.54,453.04) node  [font=\tiny,rotate=-270] [align=left] {Agent-1};
\draw (466.54,480.51) node  [font=\tiny,rotate=-270] [align=left] {Agent-2};
\draw (466.54,508.55) node  [font=\tiny,rotate=-270] [align=left] {Agent-3};
\draw (542.7,437.46) node  [font=\tiny] [align=left] {Agent-3};
\draw (514.66,437.46) node  [font=\tiny] [align=left] {Agent-2};
\draw (486.04,437.46) node  [font=\tiny] [align=left] {Agent-1};

\end{tikzpicture}\label{subfig:Hall_structure}}
\caption{An example of the $\mat{H}{}{}$ matrix structure computed with the three-agent setup shown in \figref{fig:multi_factorgraph}}
\label{fig:Hmat_structure}
\end{figure}
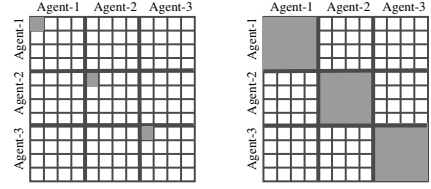
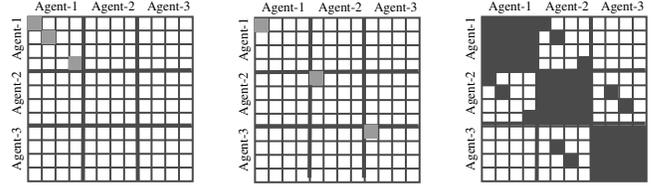

The state variables $\unknown$ can be estimated by solving the nonlinear LS problem iteratively. With an initial guess of the poses $\{\mat{S}{GC_k}{n}\}$, we use the Levenberg-Marquardt algorithm \cite{levenberg1944method, marquardt1963algorithm} to iteratively update the estimate, employing the gtsam open source library \cite{dellaert2012factor}. 
At each iteration, this algorithm estimates the pose updates which are parameterized as the $7\times1$ twist vectors (non-zero elements of the Lie algebra of $\mathrm{Sim(3)}$ group)
\begin{align}
 \Delta\unknownLieVecOpt = \{ \{\boldxiOptw{GC_k}{n_0},\dots, \boldxiOptw{GC_k}{n_N}\}, \forall k \in \{1,2,\dots,K\} \}.
\end{align}
$\Delta\unknownLieVecOpt$ can be estimated by solving the normal equation
\begin{align}
 ( \mat{H}{}{}+d\mat{I}{}{} )\Delta\unknownLieVecOpt =  -\vec{b}{}{}{}{}, \label{eq:normal_eq}
\end{align}
where $d$ is a scalar (damping factor) introduced for better convergence. $\mat{H}{}{}$ is the $7\times\sum_{k}7N_k$ matrix when $N_k$ is the total number of agent-$k$'s keyframe poses:
\begin{align}
\mat{H}{}{} = & \underbrace{ \sum_{k,\kprime}\sum_{i}\mat{H}{r,k\kprime}{i} }_\text{Inter-agent range}
            + \underbrace{ \sum_{k}\sum_{i}\mat{H}{r,k}{i} }_\text{Anchor range} \nonumber \\
            & + \underbrace{ \sum_{k}\sum_{i,j}\mat{H}{odo,k}{ij} }_\text{Odometry}
            + \underbrace{ \sum_{k}\mat{H}{pri,k}{1} }_\text{A prior}
\label{eqn:H_matrix}
\end{align}
Each $\mat{H}{*,k}{i}$ matrix in the right-side terms ($k$ and $i$ here as general agent indices and time indices respectively) of Eqn. (\ref{eqn:H_matrix}) is computed using the corresponding Jacobian $\jacobian{*,k}{i}$ and covariance matrices $\mathbf{W}_{*,k}^i$ as 
\begin{align}
   \mat{H}{*,k}{i} = \transpose{\jacobian{*,k}{i}}\mathbf{W}_{*,k}^i \jacobian{*,k}{i}.
\end{align}
Additionally, $\mathbf{b}$ on the right hand side of \eqref{eq:normal_eq} is a $\sum_{k}7N_k \times 1$ vector:
\begin{align}
  &\mathbf{b}  = \sum_{k}\vec{b}{pri,k}{1}{}  
               +\sum_{k}\sum_{i,j}\vec{b}{odo,k}{ij}{}
               +\sum_{k}\sum_{i}\vec{b}{r,k}{i}{}
               +\sum_{k,\kprime}\sum_{i}\vec{b}{r,k\kprime}{i}{} 
\end{align}
where each $\mathbf{b}_{*,k}^{i}$ vector ($k$ and $i$ here as general agent indices and time indices respectively) is computed using the corresponding Jacobian matrices and the residuals in the measurements using the estimated states from last iteration $\Delta \mathbf{z}_{*,k}^i$:
\begin{align}
   \vec{b}{*,k}{i}{} = \transpose{\jacobian{*,k}{i}}\mathbf{W}_{*,k}^i \Delta \mathbf{z}_{*,k}^i.
\end{align}

\figref{fig:Hmat_structure} shows the structure of the $\mat{H}{}{}$ matrix computed using the three-agent setup depicted in \figref{fig:multi_factorgraph}. A small box in the figure represents the 7DoF pose at each timestamp. Only the non-zeros elements of the matrix are marked in gray. As shown in this figure, both agent-to-anchor and inter-agent range measurements do not induce a big complexity to the system, so we can carry out the data fusion without requiring substantial computational power.

In each iteration, the estimated twist vector $\Delta\unknownLieVecOpt$ using \eqrefnew{eq:normal_eq} is converted back to the matrix domain, i.e. $\{\Delta\matOpt{S}{GC_k}{n}\}$. Then, the robots' poses are updated with the optimal solutions $\Delta\matOpt{S}{GC_k}{n}$ until the Levenberg-Marquardt algorithm converges:
\begin{align}
   {\matOpt{S}{GC_k}{n}} \leftarrow \mat{S}{GC_k}{n} \oplus \Delta\matOpt{S}{GC_k}{n}, 
\end{align}
where $\oplus$ is the multiplication of two matrices in the Lie group. Detail explanation of the operation can be found in \cite{hall2015lie}. 
%
The pose estimation can be calculated incrementally by applying the iSAM method \cite{kaess2012isam2}.

The LS estimator in Eqn. (\ref{eqn:SSE_measurements}) is optimal in the sense of MAP when the measurement noise is zero mean Gaussian. Although measurement errors in practice contains biases in many cases, the estimator is still effective as long as the system have sufficient redundancy, which can be reflected from our tests in Section \ref{sec:single_exp} and \ref{sec:system_outdoor_test} using realistic ranging error models.

After the multi-agent data fusion, each robot receives the updated keyframe poses through the communication links, and then the local map points are updated on each robot's onboard computer using the differences of the keyframe poses before and after the data fusion.

\section{System test using experimental data}
\label{sec:single_exp}

\begin{figure}[t!]
\centering
\captionsetup[subfloat]{labelfont=normalsize,textfont=normalsize}
\subfloat[Rover-1]{\centering\includegraphics[width=.45\linewidth]{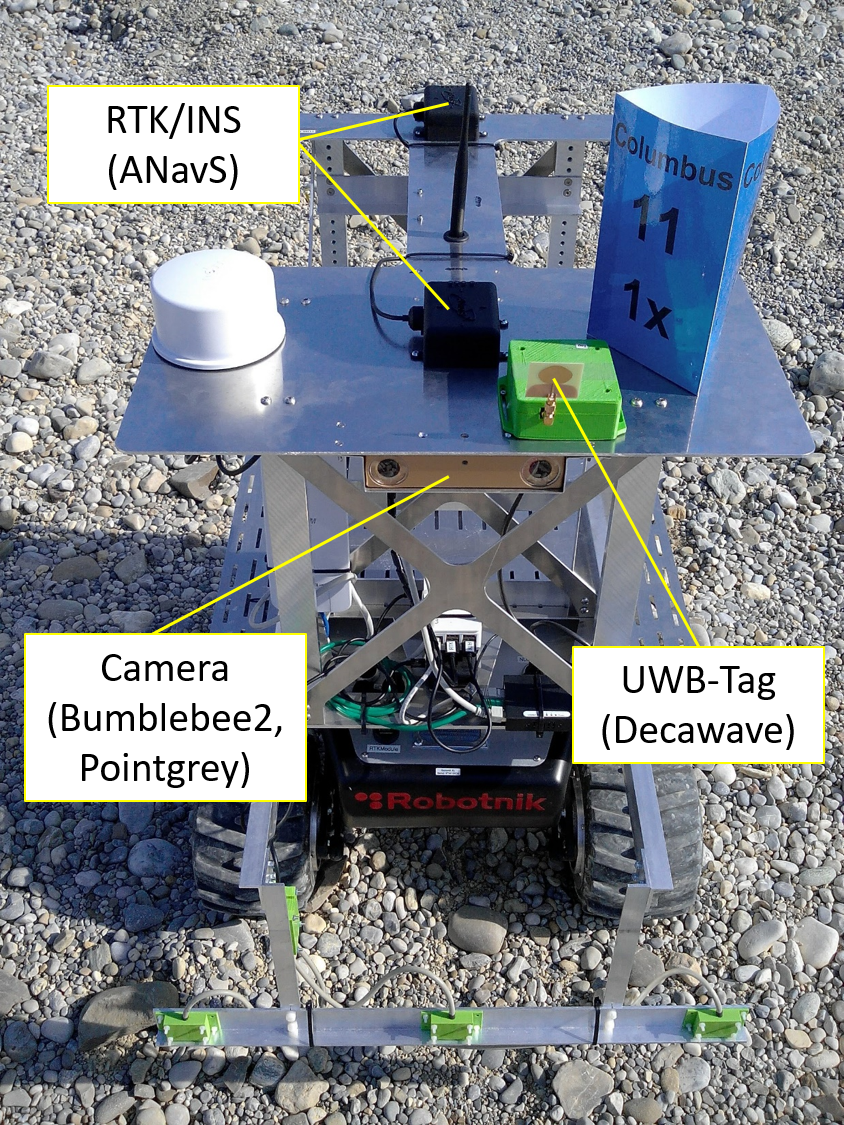} \label{subfig:singleExp_dlr_dynamic_rover}} ~~~
\subfloat[Rover-2]{\centering\includegraphics[width=.45\linewidth]{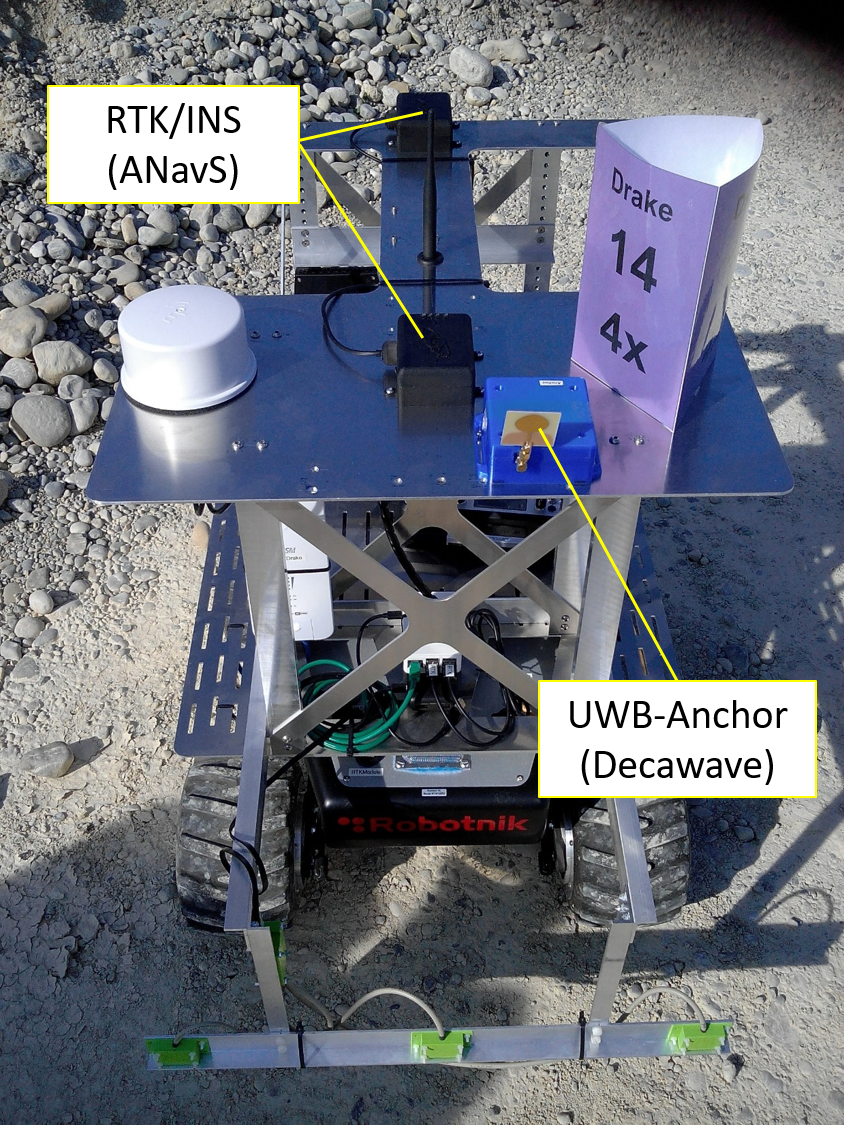} \label{subfig:singleExp_dlr_static_rover}} 
\caption{The test platforms developed by the Institute of Communications and Navigation, Germany Aerospace Center (DLR-KN)}
\label{fig:singleExp_dlr_rover_system}
\end{figure}

\begin{figure}[t!]
\centering
\captionsetup[subfloat]{labelfont=normalsize,textfont=normalsize}
\subfloat[Football field]{\centering\includegraphics[width=.45\linewidth]{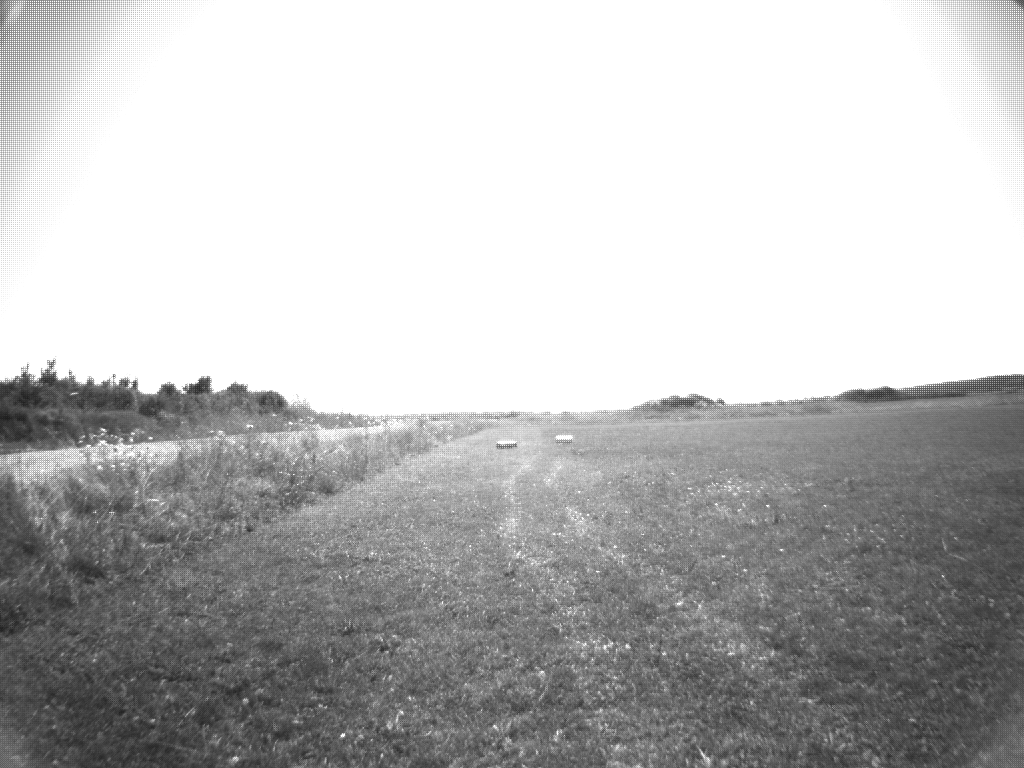}\label{subfig:singleExp_imgSample_football}} ~~~
\subfloat[Gravel pit]{\centering\includegraphics[width=.45\linewidth]{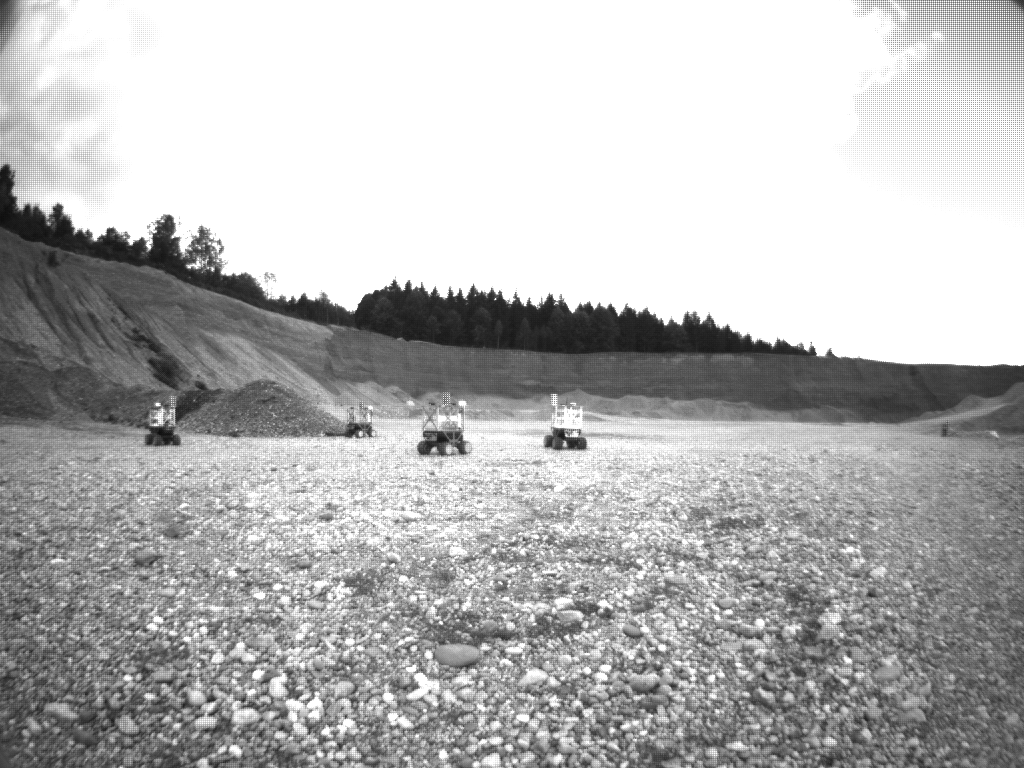}\label{subfig:singleExp_imgSample_kiesgrube}}
\caption{Image samples of a football field in Oberpfaffenhofen and a gravel pit in Planegg, Germany}
\label{fig:singleExp_imgSample}
\end{figure}

Using experimental data, we show in this section that CoVOR-SLAM can significantly reduce the estimation error of visual odometry without the scale ambiguity. Moreover, CoVOR-SLAM can achieve similar positioning accuracy as the one of the inter-agent loop closures, with a much reduced computational complexity.

\subsection{Experimental setup}
Experimental data was obtained on a football field and in a gravel pit using two rovers developed by the Institute of Communications and Navigation, Germany Aerospace Center (DLR-KN). In this experiment, Ultra-Wide Band (UWB) sensors are used to obtain ranging measurements.
Rover-1 (see \figref{subfig:singleExp_dlr_dynamic_rover}) has a front-looking camera (Bumblebee2, pointgrey, 20fps) and an IEEE 802.15.4-2011 UWB standard ranging sensor with DW1000 IC chips from Decawave (Qorvo), as well as GNSS Real-Time Kinematic (RTK) and Inertial Navigation Systems (INS) to obtain the ground truth poses. \figref{fig:singleExp_imgSample} shows the sample images obtained using the left camera in the test environments.  
Rover-2 has an UWB ranging sensor and stayed at a fixed location as an anchor station (see \figref{subfig:singleExp_dlr_static_rover}).

\begin{figure}[t!]
\centering
\captionsetup[subfloat]{labelfont=normalsize,textfont=normalsize}
\subfloat[Football field]{\centering\includegraphics[width=.45\linewidth]{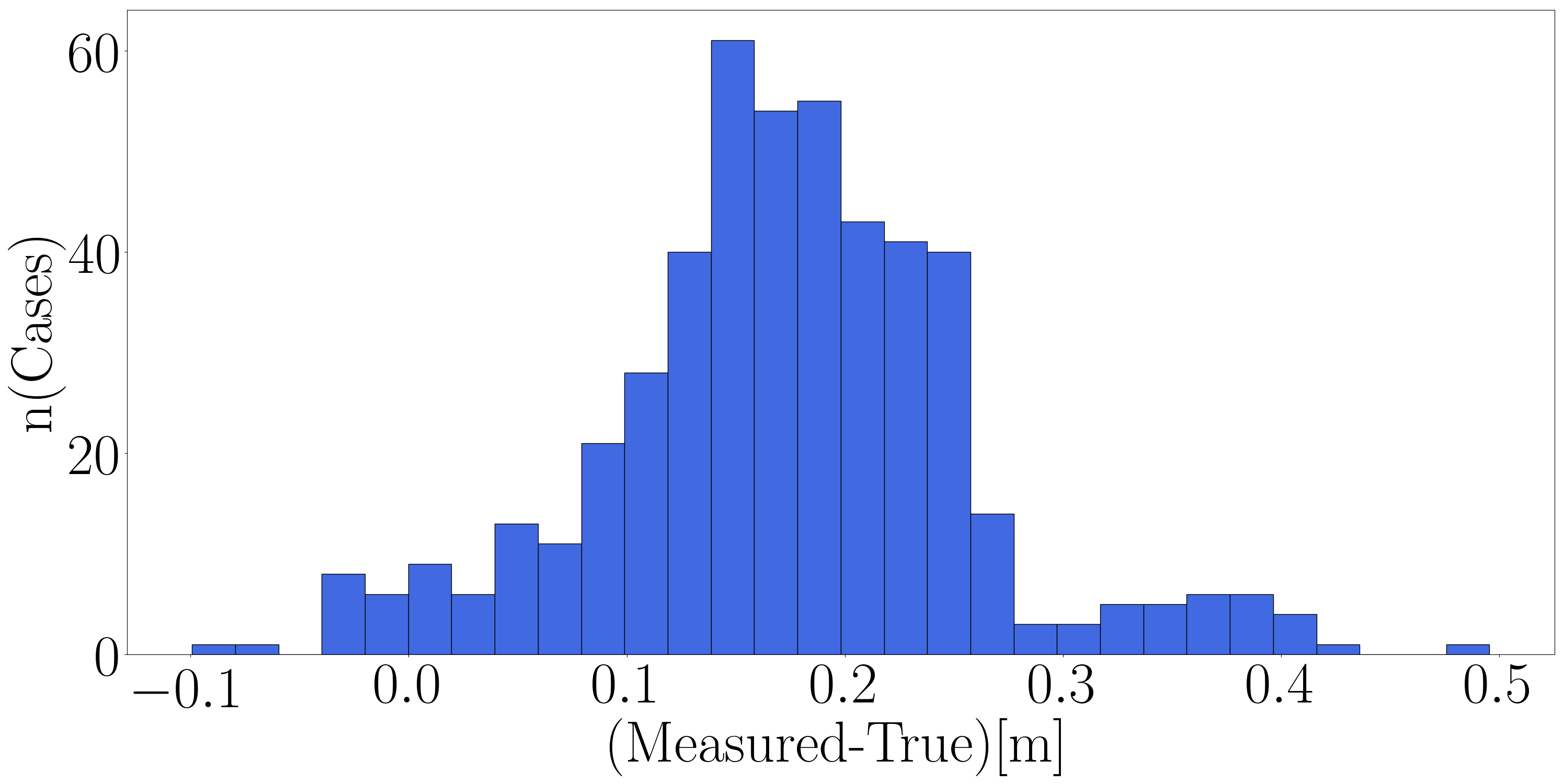} \label{subfig:singleExp_rangeMeasErrHist_football}} ~~
\subfloat[Gravel pit]{\centering\includegraphics[width=.45\linewidth]{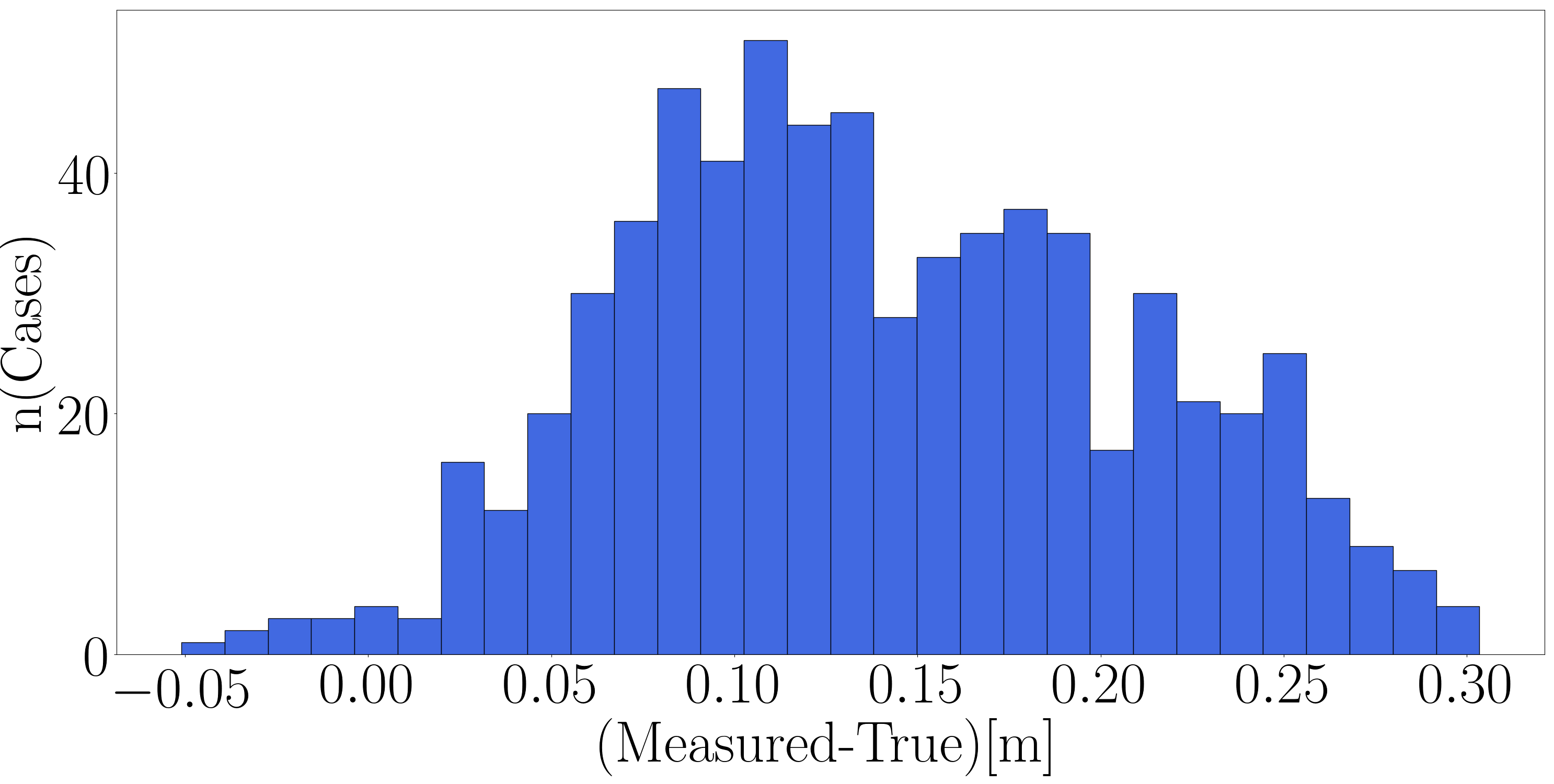} \label{subfig:singleExp_rangeMeasErrHist_kiesgrube}} 
\caption{The histograms of the ranging errors obtained in two outdoor environments }
\label{fig:singleExp_rangeMeasErrHist}
\end{figure}

\figref{fig:singleExp_rangeMeasErrHist} shows the ranges between the UWB modules of Rover-1 and Rover-2. As shown in this figure, the UWB range measurements are biased, and not simply zero-mean Gaussian distributed. 
\tabref{table:singleExp_summary_football_and_kiesgrube} summarizes the the datasets' total traveling time, length, and the dimension of the trajectories. 

\begin{table}[t!]
 \centering
 \caption{Summary of the football field and gravel pit datasets}
 \label{table:singleExp_summary_football_and_kiesgrube}
 \resizebox{\columnwidth}{!}{
 \begin{tabular}[c]{|c|c|c|c|}
  \hline
  Dataset & \makecell{Total time \\ $[\snd]$} & \makecell{Total length \\ $[\m]$ } & \makecell{Dimension \\ ($\mathrm{X}[\m]\times\mathrm{Y}[\m]\times\mathrm{Z}[\m]$)}\\
  \hline\hline
  Football field & 494.01& 105.61 & $25.54\times26.25\times0.18$ \\
  Gravel pit     & 457.17& 38.98  & $8.05\times9.29\times0.17$ \\
  \hline
 \end{tabular}}
\end{table}

\subsection{Analysis of the positioning accuracy}
\begin{figure}[t!]
\centering
\captionsetup[subfloat]{labelfont=normalsize,textfont=normalsize}
\subfloat[Football field]{\centering\includegraphics[width=.95\linewidth]{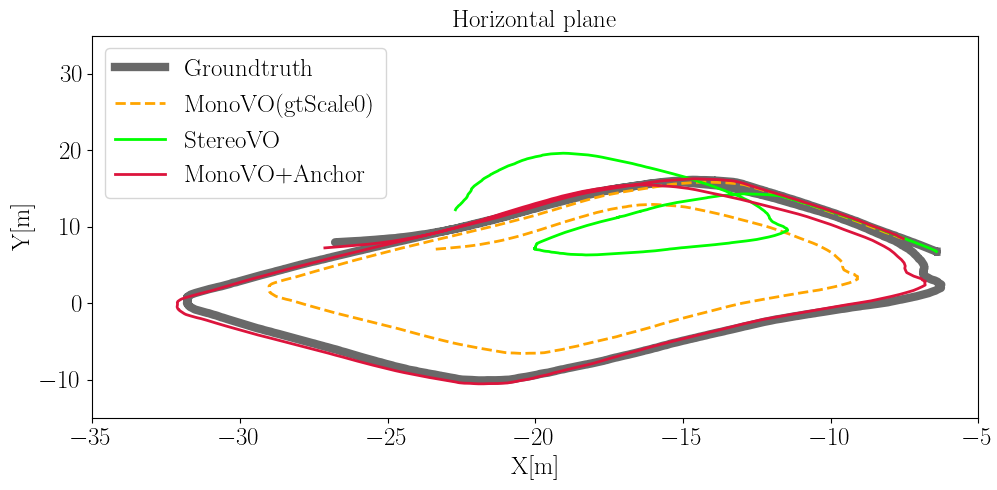} \label{subfig:singleExp_hori_traj_football}} \\
\subfloat[Gravel pit]{\centering\includegraphics[width=.95\linewidth]{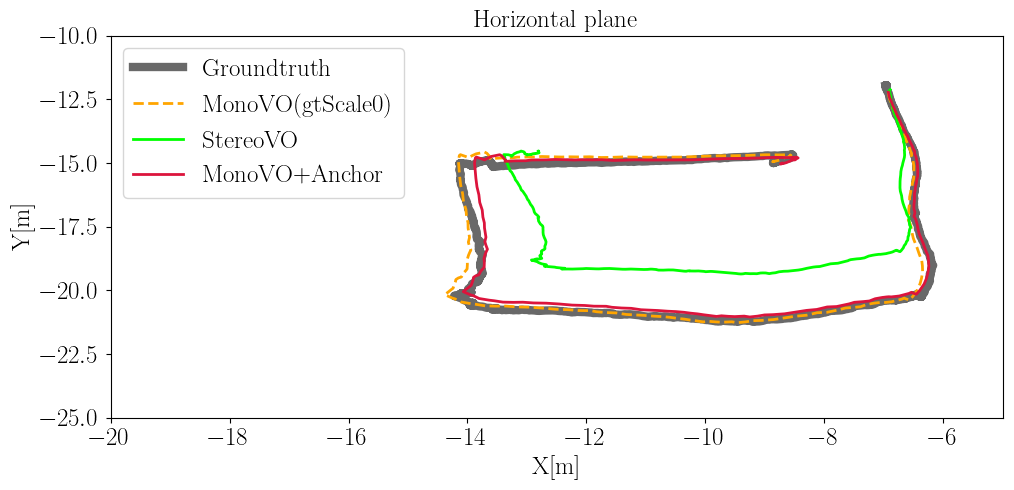} \label{subfig:singleExp_hori_traj_kiesgrube}} 
\caption{The ground truth is depicted as the gray solid lines. Horizontal trajectories estimated using monocular VO and scaled with the ground truth value (orange), stereo VO (green), and CoVOR-SLAM (red). }
\label{fig:singleExp_hori_traj}
\end{figure}

First, trajectories of Rover-1 were estimated using monocular VO (orange lines in \figref{fig:singleExp_hori_traj}). The source codes of ORB-SLAM \cite{mur2017orb} was used to process monocular VO.  
As shown in \figref{fig:singleExp_hori_traj}, the shape of the orange lines are similar to the ground truth trajectories (gray lines). 
However, the trajectories estimated using monocular VO are up-to-scale, so the absolute scale was manually determined using the ground truth value (obtained with GNSS and INS) at the beginning of the mission.

CoVOR-SLAM was used to resolve the scale ambiguity of monocular VO and to mitigate the pose estimation errors. The red solid lines in \figref{fig:singleExp_hori_traj} show the trajectories estimated using CoVOR-SLAM with monocular VO and UWB range measurements. 
As described in \secref{subsec:multi_agent_data_fusion}, robot's poses were estimated as the 7DoF variables in the framework, i.e. the  scale value of the translation were estimated at each timestamp using the information from the sparse range measurements.  

The figure shows that the trajectories were accurately estimated using CoVOR-SLAM with real range measurements, the error of which are in general biased and not necessarily Gaussian distributed (see \figref{fig:singleExp_rangeMeasErrHist}). This result shows that CoVOR-SLAM is robust to the range measurement biases.

Additionally, we estimated the rover's trajectories using stereo images without range measurements (green lines in \figref{fig:singleExp_hori_traj}). The absolute scales were inaccurately estimated so that the estimated poses drift away from the ground truth fast for both experiments, because the baseline length of the camera was too short (only $\sim$12cm) to accurately estimate the depth, which is a common setup for a stereo rig on small robots. 

Furthermore, \figref{subfig:singleExp_hori_traj_football} shows that the shape of the trajectory estimated using stereo visual odometry was completely different from the ground truth for the football field dataset, i.e. the translation vectors were also inaccurately estimated using stereo VO. 
This issue could be induced by feature mismatch. On the football field, features were detected mostly in grass, it is challenging to differentiate them (since their descriptors are very similar). When similar features are observed by two cameras in a stereo rig with short baseline length, they are located close to each other in different images, which makes them even more challenging to be differentiated. 

On the contrary, CoVOR-SLAM can accurately estimate the translation vectors even in such environments, using dynamic scene changes of monocular VO, and resolve the scale ambiguity of monocular VO, exploiting ranges between the rovers. As a consequence, the red line almost perfectly follows the ground truth trajectory (gray line) in \figref{subfig:singleExp_hori_traj_football}.

\subsection{Comparison to the loop closures}

We also compare CoVOR-SLAM to the loop closures in terms of positioning accuracy and computational requirement. Only the football field dataset was used, because the rover did not revisit any known location in the gravel pit dataset so that loops were not detected. 

\begin{figure}[t!]  
  \centering
  \includegraphics[width=.95\linewidth]{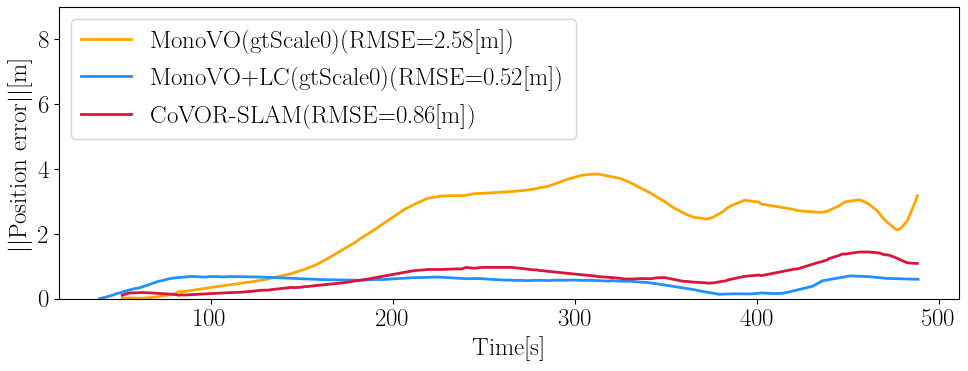}
  \caption{The magnitude of the difference between the ground truth trajectory and position vectors estimated using monocular visual odometry (orange), monocular visual odometry and loop closing (blue), and CoVOR-SLAM (red) }
  \label{fig:singleExp_pos_rmse_vsLC_football}
\end{figure}

\figref{fig:singleExp_pos_rmse_vsLC_football} shows the differences between the ground truth trajectory and positions estimated using monocular visual odometry (orange), monocular visual odometry and loop closing (blue), and CoVOR-SLAM (red). We can see that the RMSE of position estimates was smallest when the loop closing technique was used. However, the loop closing technique requires the rover to revisit known locations to mitigate the drift. Moreover, the absolute scale cannot be estimated using the monocular camera alone, so the trajectory estimate was manually scaled using ground truth in the plot. 
On the contrary, CoVOR-SLAM achieved the similar RMSE of positioning as the one obtained with loop closures, without manually scaling the trajectory, i.e. CoVOR-SLAM accurately estimated the positions as well as the absolute scale without any constraint on the motion trajectory of the robots.

\begin{table}[t!]
 \centering
 \caption{Processing time per keyframe [ms] - Football field}
 \label{table:procTime_per_kf_football}
 \begin{tabular}[c]{|cc|cc|}
  \hline
  \multicolumn{2}{|c|}{Loop Closing}& \multicolumn{2}{c|}{CoVOR-SLAM (Ours)}\\
  \hline
  DetectLoop & CorrectLoop & CheckFusion & DataFusion \\
  \hline
   0.088 & 14.730  & $3.520\times10^{-7}$ & 0.860 \\
  \hline
 \end{tabular}
\end{table}

Furthermore, \tabref{table:procTime_per_kf_football} shows the time required to process the most computationally demanding parts of loop closing and CoVOR-SLAM. A laptop with Intel® Core™ i7-7700HQ CPU @ 2.80GHz $\times$ 8, Ubuntu 20.04.2 LTS was used for data processing. As shown in this table, it took $0.088\ms$ per keyframe to search loop candidates, while CoVOR-SLAM requires almost no time to decide whether we need to process the data fusion or not (CheckFusion). It took $14.730\ms$ per keyframe to correct the graph using the loop information (CorrectLoop), while it only took $0.860\m$ per keyframe to fuse the visual odometry and range measurements (DataFusion), which is about $94.2\%$ less than the loop closing. The computational cost advantage of CoVOR-SLAM becomes even more significantly as the number of agents increases, when the robot swarm needs to detect inter-agent loops.

Additionally, CoVOR-SLAM can be used for both datasets, while loop closing was only available for the football field dataset, although rover-1 came back very close to its starting point at the end of the missions for the gravel pit dataset.

\section{System analysis with public datasets employing four-agent setup}
\label{sec:system_outdoor_test}

\begin{figure}[t!]  
  \centering
  \includegraphics[width=.95\linewidth]{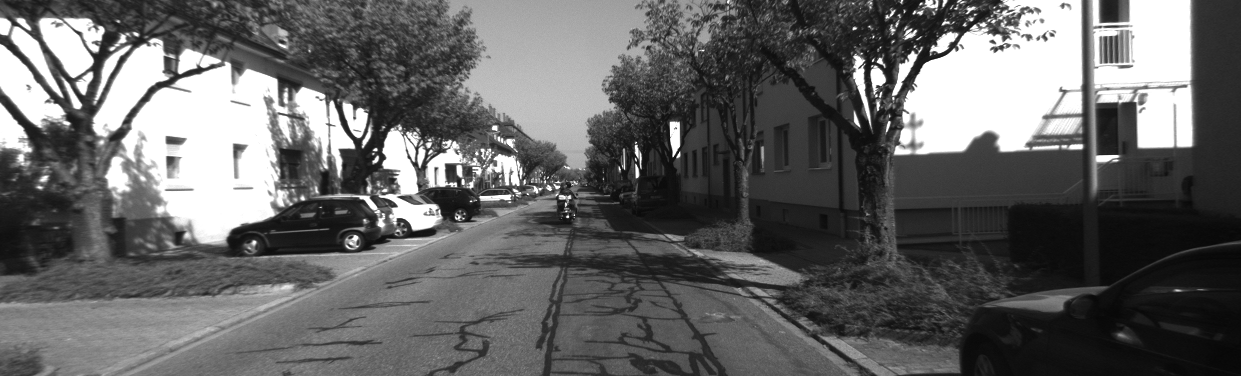}
  \caption{A sample image of the KITTI dataset sequence 00}
  \label{fig:multiEval_imgSample_kitti_00}
\end{figure}

\begin{figure}[t!]  
  \centering
  \includegraphics[width=.95\linewidth]{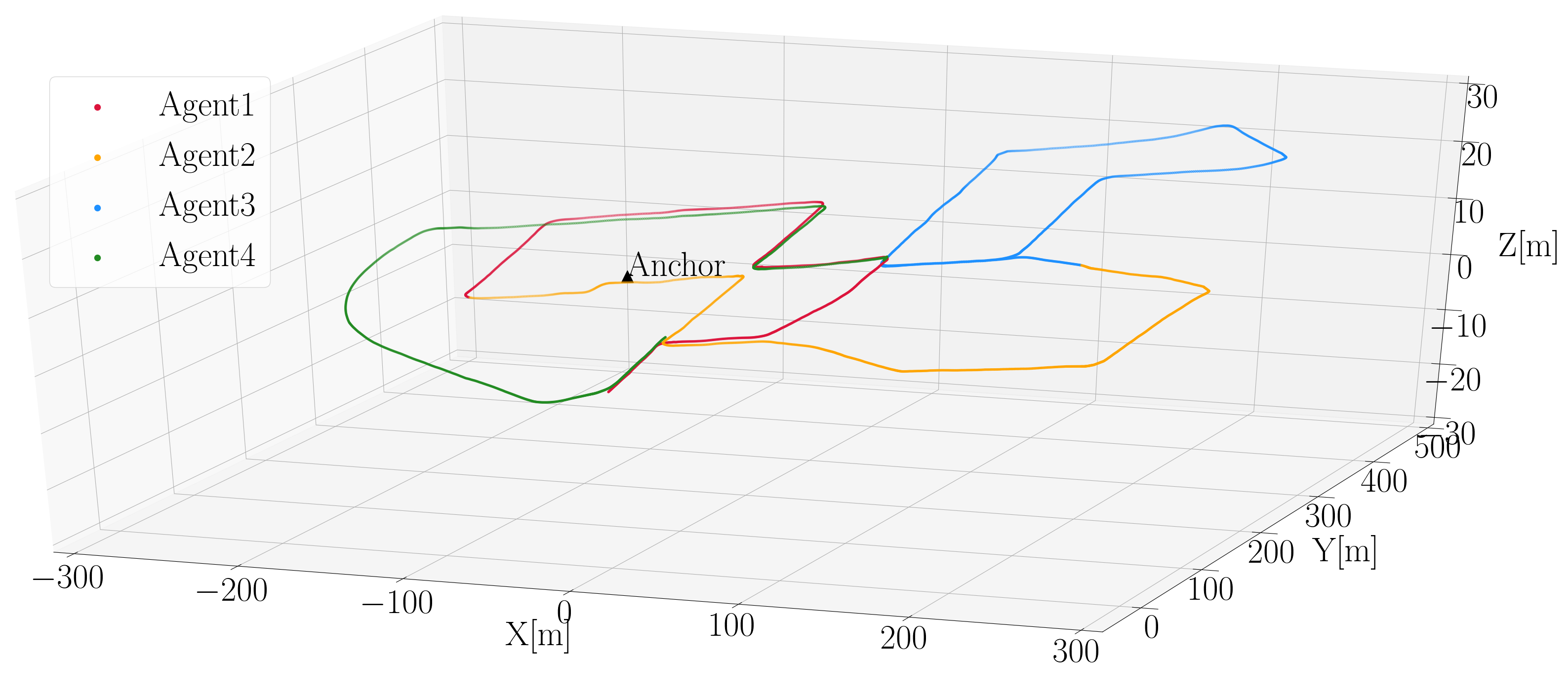}
  \caption{Ground truth trajectories of four agents. The anchor is located at $[-50, 150, 10]\m^\mathrm{i}$}
  \label{fig:multiEval_traj_gt_kitti_00}
\end{figure}

CoVOR-SLAM has also been tested using a larger multi-agent setup, also taking into account the effect of connectivity between robots. The sequence 00 of the KITTI dataset \cite{geiger2012we} was used for the test. The original images were obtained using a single car in a residential area (see \figref{fig:multiEval_imgSample_kitti_00}). We divided the dataset into four segments, assuming that four robot agents simultaneously traveled each part. \figref{fig:multiEval_traj_gt_kitti_00} shows the ground truth trajectory of each agent and the position of a single anchor station. 

\subsection{Ranges including errors modeled using UWB measurements}
\label{subsec:ranging_err_modeling}

Since the public dataset does not contain ranges, we generated range measurements using the error model created with the UWB range measurements (see \figref{fig:singleExp_rangeMeasErrHist}). This error model includes zero-mean Gaussian noise $\eta$ as well as the systematic errors $b_{sys}$ and multipath errors $b_{m}$: 
\begin{align}
   \varepsilon = b_{sys} + b_{m} + \eta . \label{eq:singleExp_ranging_err_model}
\end{align}

\begin{figure}[t!]
 \centering
\tikzset{every picture/.style={line width=0.75pt}} 

\begin{tikzpicture}[x=0.75pt,y=0.75pt,yscale=-1,xscale=1]

\draw [color={rgb, 255:red, 74; green, 144; blue, 226 }  ,draw opacity=1 ] [dash pattern={on 4.5pt off 4.5pt}]  (216.14,59.63) -- (63.21,67.82) ;
\draw  [draw opacity=0] (54.02,64.89) .. controls (55.74,63.41) and (58.09,62.07) .. (60.81,61.1) .. controls (63.17,60.27) and (65.48,59.84) .. (67.52,59.77) -- (62.83,66.93) -- cycle ; \draw   (54.02,64.89) .. controls (55.74,63.41) and (58.09,62.07) .. (60.81,61.1) .. controls (63.17,60.27) and (65.48,59.84) .. (67.52,59.77) ;  
\draw  [draw opacity=0] (57.88,66.57) .. controls (59.07,65.82) and (60.45,65.13) .. (61.97,64.57) .. controls (63.5,64) and (65.01,63.63) .. (66.42,63.43) -- (63.67,69.2) -- cycle ; \draw   (57.88,66.57) .. controls (59.07,65.82) and (60.45,65.13) .. (61.97,64.57) .. controls (63.5,64) and (65.01,63.63) .. (66.42,63.43) ;  
\draw  [draw opacity=0] (51.8,61.51) .. controls (53.87,59.97) and (56.44,58.58) .. (59.35,57.5) .. controls (62.44,56.36) and (65.49,55.73) .. (68.19,55.6) -- (62.13,65.04) -- cycle ; \draw   (51.8,61.51) .. controls (53.87,59.97) and (56.44,58.58) .. (59.35,57.5) .. controls (62.44,56.36) and (65.49,55.73) .. (68.19,55.6) ;  

\draw   (82.09,84.06) .. controls (83.37,83.58) and (84.83,84.22) .. (85.35,85.48) -- (86.11,87.31) .. controls (86.63,88.58) and (86.01,89.98) .. (84.73,90.46) -- (84.73,90.46) .. controls (83.45,90.93) and (81.99,90.29) .. (81.47,89.03) -- (80.71,87.2) .. controls (80.19,85.94) and (80.81,84.53) .. (82.09,84.06) -- cycle ;
\draw   (76.67,70.95) .. controls (77.95,70.48) and (79.41,71.11) .. (79.93,72.37) -- (80.69,74.21) .. controls (81.21,75.47) and (80.59,76.88) .. (79.32,77.35) -- (79.32,77.35) .. controls (78.04,77.83) and (76.58,77.19) .. (76.05,75.93) -- (75.3,74.09) .. controls (74.78,72.83) and (75.39,71.43) .. (76.67,70.95) -- cycle ;
\draw   (60.3,92.14) .. controls (61.58,91.66) and (63.04,92.3) .. (63.56,93.56) -- (64.32,95.4) .. controls (64.84,96.66) and (64.23,98.06) .. (62.95,98.54) -- (62.95,98.54) .. controls (61.67,99.01) and (60.21,98.38) .. (59.69,97.12) -- (58.93,95.28) .. controls (58.41,94.02) and (59.02,92.61) .. (60.3,92.14) -- cycle ;
\draw   (54.88,79.03) .. controls (56.16,78.56) and (57.62,79.2) .. (58.15,80.46) -- (58.9,82.29) .. controls (59.42,83.55) and (58.81,84.96) .. (57.53,85.44) -- (57.53,85.44) .. controls (56.25,85.91) and (54.79,85.27) .. (54.27,84.01) -- (53.51,82.18) .. controls (52.99,80.92) and (53.61,79.51) .. (54.88,79.03) -- cycle ;
\draw  [fill={rgb, 255:red, 255; green, 255; blue, 255 }  ,fill opacity=1 ] (70.53,71.36) .. controls (72.58,70.6) and (74.91,71.62) .. (75.75,73.64) -- (82.1,89) .. controls (82.93,91.02) and (81.95,93.27) .. (79.9,94.03) -- (68.78,98.16) .. controls (66.73,98.92) and (64.39,97.9) .. (63.56,95.88) -- (57.21,80.51) .. controls (56.38,78.5) and (57.36,76.25) .. (59.41,75.49) -- cycle ;

\draw   (62.38,69.71) .. controls (62.08,68.96) and (62.46,68.11) .. (63.21,67.82) .. controls (63.96,67.52) and (64.81,67.9) .. (65.1,68.65) .. controls (65.39,69.4) and (65.02,70.25) .. (64.27,70.54) .. controls (63.52,70.83) and (62.67,70.46) .. (62.38,69.71) -- cycle ;
\draw  [draw opacity=0] (210.75,49.94) .. controls (213.02,49.92) and (215.68,50.39) .. (218.39,51.4) .. controls (220.73,52.28) and (222.78,53.43) .. (224.39,54.7) -- (216.19,57.16) -- cycle ; \draw   (210.75,49.94) .. controls (213.02,49.92) and (215.68,50.39) .. (218.39,51.4) .. controls (220.73,52.28) and (222.78,53.43) .. (224.39,54.7) ;  
\draw  [draw opacity=0] (212.63,53.71) .. controls (214.02,53.9) and (215.52,54.26) .. (217.05,54.8) .. controls (218.59,55.35) and (219.99,56.04) .. (221.19,56.79) -- (215.38,59.45) -- cycle ; \draw   (212.63,53.71) .. controls (214.02,53.9) and (215.52,54.26) .. (217.05,54.8) .. controls (218.59,55.35) and (219.99,56.04) .. (221.19,56.79) ;  
\draw  [draw opacity=0] (211.22,45.92) .. controls (213.8,46.07) and (216.67,46.66) .. (219.58,47.7) .. controls (222.69,48.81) and (225.42,50.29) .. (227.58,51.93) -- (216.87,55.27) -- cycle ; \draw   (211.22,45.92) .. controls (213.8,46.07) and (216.67,46.66) .. (219.58,47.7) .. controls (222.69,48.81) and (225.42,50.29) .. (227.58,51.93) ;  

\draw   (219.94,82.66) .. controls (221.23,83.12) and (221.94,84.54) .. (221.53,85.85) -- (220.93,87.74) .. controls (220.52,89.04) and (219.14,89.72) .. (217.86,89.26) -- (217.86,89.26) .. controls (216.57,88.81) and (215.86,87.38) .. (216.27,86.08) -- (216.87,84.18) .. controls (217.28,82.88) and (218.66,82.2) .. (219.94,82.66) -- cycle ;
\draw   (224.21,69.14) .. controls (225.5,69.6) and (226.21,71.02) .. (225.8,72.32) -- (225.2,74.22) .. controls (224.79,75.52) and (223.41,76.2) .. (222.13,75.74) -- (222.13,75.74) .. controls (220.84,75.29) and (220.13,73.86) .. (220.54,72.56) -- (221.14,70.66) .. controls (221.55,69.36) and (222.93,68.68) .. (224.21,69.14) -- cycle ;
\draw   (198.06,74.86) .. controls (199.34,75.31) and (200.05,76.74) .. (199.64,78.04) -- (199.04,79.94) .. controls (198.63,81.24) and (197.26,81.92) .. (195.97,81.46) -- (195.97,81.46) .. controls (194.69,81) and (193.98,79.58) .. (194.39,78.28) -- (194.99,76.38) .. controls (195.4,75.08) and (196.77,74.4) .. (198.06,74.86) -- cycle ;
\draw   (202.33,61.33) .. controls (203.61,61.79) and (204.32,63.22) .. (203.91,64.52) -- (203.31,66.41) .. controls (202.9,67.72) and (201.53,68.4) .. (200.24,67.94) -- (200.24,67.94) .. controls (198.95,67.48) and (198.25,66.05) .. (198.66,64.75) -- (199.26,62.86) .. controls (199.67,61.56) and (201.04,60.88) .. (202.33,61.33) -- cycle ;
\draw  [fill={rgb, 255:red, 255; green, 255; blue, 255 }  ,fill opacity=1 ] (219.25,65.51) .. controls (221.3,66.24) and (222.44,68.52) .. (221.78,70.61) -- (216.77,86.46) .. controls (216.12,88.54) and (213.91,89.63) .. (211.86,88.9) -- (200.68,84.91) .. controls (198.63,84.18) and (197.49,81.9) .. (198.15,79.82) -- (203.16,63.96) .. controls (203.82,61.88) and (206.02,60.79) .. (208.07,61.52) -- cycle ;

\draw   (214.06,59) .. controls (214.32,58.24) and (215.15,57.83) .. (215.91,58.09) .. controls (216.68,58.35) and (217.09,59.18) .. (216.83,59.94) .. controls (216.57,60.7) and (215.74,61.11) .. (214.98,60.86) .. controls (214.21,60.6) and (213.8,59.77) .. (214.06,59) -- cycle ;
\draw [color={rgb, 255:red, 74; green, 144; blue, 226 }  ,draw opacity=1 ]   (64.27,70.54) -- (51.19,36.67) ;
\draw [shift={(50.47,34.8)}, rotate = 68.88] [color={rgb, 255:red, 74; green, 144; blue, 226 }  ,draw opacity=1 ][line width=0.75]    (10.93,-3.29) .. controls (6.95,-1.4) and (3.31,-0.3) .. (0,0) .. controls (3.31,0.3) and (6.95,1.4) .. (10.93,3.29)   ;
\draw [color={rgb, 255:red, 74; green, 144; blue, 226 }  ,draw opacity=1 ]   (215.91,58.09) -- (226.39,28.44) ;
\draw [shift={(227.06,26.55)}, rotate = 109.46] [color={rgb, 255:red, 74; green, 144; blue, 226 }  ,draw opacity=1 ][line width=0.75]    (10.93,-3.29) .. controls (6.95,-1.4) and (3.31,-0.3) .. (0,0) .. controls (3.31,0.3) and (6.95,1.4) .. (10.93,3.29)   ;
\draw  [draw opacity=0][dash pattern={on 4.5pt off 4.5pt}] (199.72,60.25) .. controls (200.03,55.95) and (201.98,51.64) .. (205.48,48.33) .. controls (209.4,44.63) and (214.44,42.95) .. (219.21,43.31) -- (216.67,60.18) -- cycle ; \draw  [color={rgb, 255:red, 74; green, 144; blue, 226 }  ,draw opacity=1 ][dash pattern={on 4.5pt off 4.5pt}] (199.72,60.25) .. controls (200.03,55.95) and (201.98,51.64) .. (205.48,48.33) .. controls (209.4,44.63) and (214.44,42.95) .. (219.21,43.31) ;  
\draw  [draw opacity=0][dash pattern={on 4.5pt off 4.5pt}] (57.4,49.31) .. controls (64.51,47.08) and (73,49.24) .. (78.94,55.48) .. controls (82.06,58.76) and (84.02,62.69) .. (84.8,66.7) -- (65.1,68.65) -- cycle ; \draw  [color={rgb, 255:red, 74; green, 144; blue, 226 }  ,draw opacity=1 ][dash pattern={on 4.5pt off 4.5pt}] (57.4,49.31) .. controls (64.51,47.08) and (73,49.24) .. (78.94,55.48) .. controls (82.06,58.76) and (84.02,62.69) .. (84.8,66.7) ;  

\draw (139.23,73.6) node    {$d$};
\draw (188.94,46.59) node    {$\theta _{2}$};
\draw (92.94,50.59) node    {$\theta _{1}$};

\end{tikzpicture}
 \caption{The definition of the angles between the LoS vector and two onboard antennas $\theta_1$, $\theta_2$, and the distance between them $d$}
 \label{fig:rangingErr_def_th_d}
\end{figure}
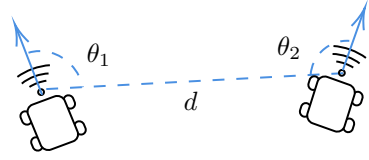

The systematic bias $b_{sys}$ is mainly induced by the signal power changes, according to the application note provided by the manufacturing company \cite{decawave2014aps011}. Thus, we modeled the systematic bias as a function of the angles between the Line-of-Sight (LoS) vector and the two ranging sensors ($\theta_0$ and $\theta_1$ in \figref{fig:rangingErr_def_th_d}) as well as the true distances between two sensors ($d$ in \figref{fig:rangingErr_def_th_d}), assuming that the signal power is mainly dependent on those three parameters. 
Additionally, the multipath bias $b_{m}$ was modeled as a random walk whose differences are Gaussian distributed. Nominal ranging noise $n$ was modeled as a zero-mean Gaussian distribution. 

\begin{figure}[t!]
   \centering
   \includegraphics[width=0.95\linewidth]{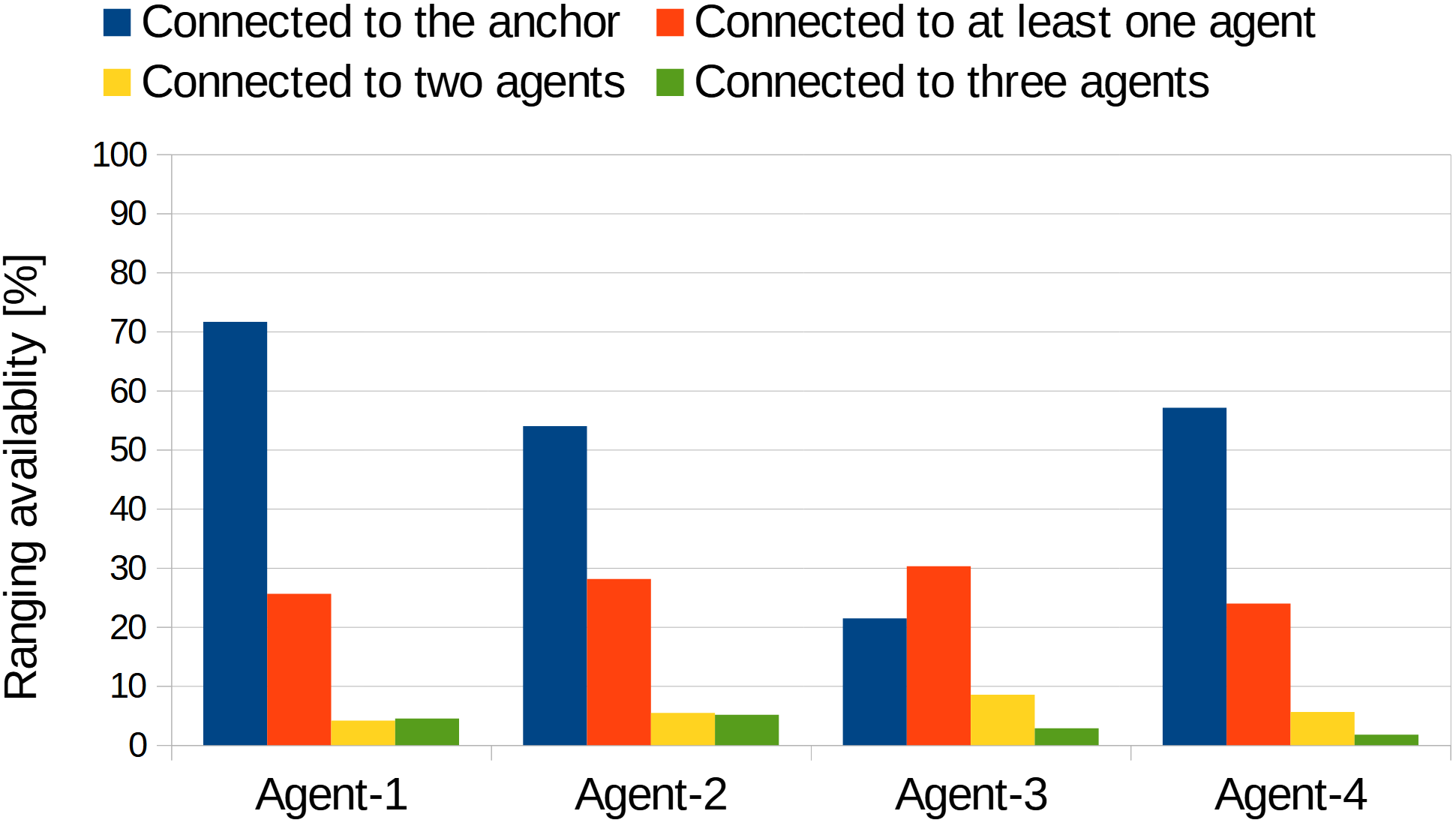}
   \caption{The range measurement availability of KITTI-00} 
   \label{fig:multiEval_summary_rangeAvail_kitti_00}
\end{figure}

Moreover, we set the maximum of ranging as $200\m$, i.e. we assume that range measurement was unavailable when the distances between two ranging modules are greater than $200\m$. \figref{fig:multiEval_summary_rangeAvail_kitti_00} shows the ranging availability of each agent compared to the total traveling time. For instance, Agent-1 obtained range measurements to the anchor point for $71.63\%$ of its total traveling time, and it was connected to at least one robot for $25.61\%$ of the mission. For $4.15\%$ and $4.50\%$ of the traveling time, it was connected to two and three other robots, respectively.

\subsection{Analysis of positioning accuracy} 

\begin{figure}[t!]
\centering
\captionsetup[subfloat]{labelfont=normalsize,textfont=normalsize}
\subfloat[Rover-1]{\includegraphics[width=.48\linewidth]{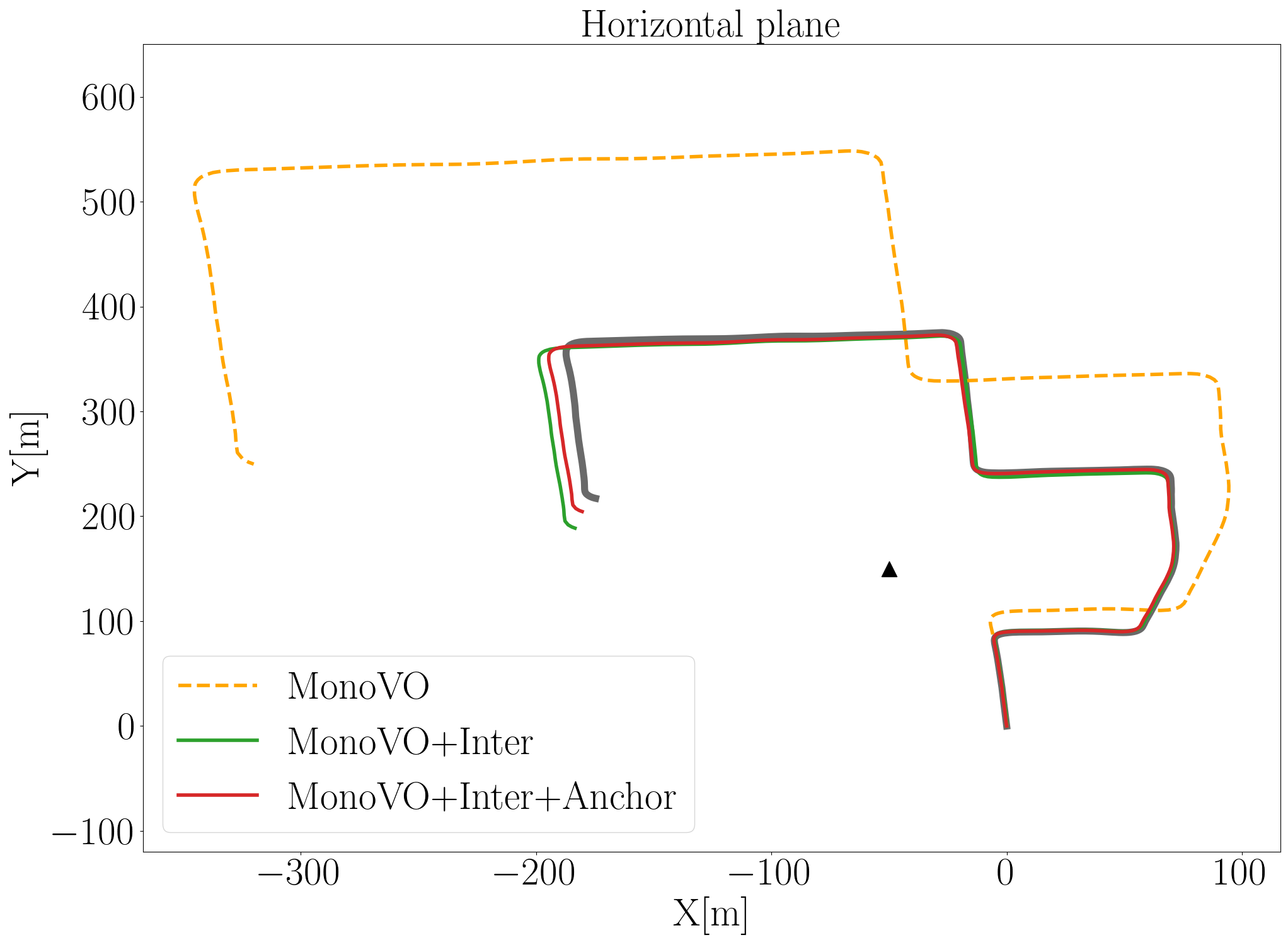}\label{subfig:multiEval_traj_est_agent1_kitti_00}} ~
\subfloat[Rover-2]{\includegraphics[width=.48\linewidth]{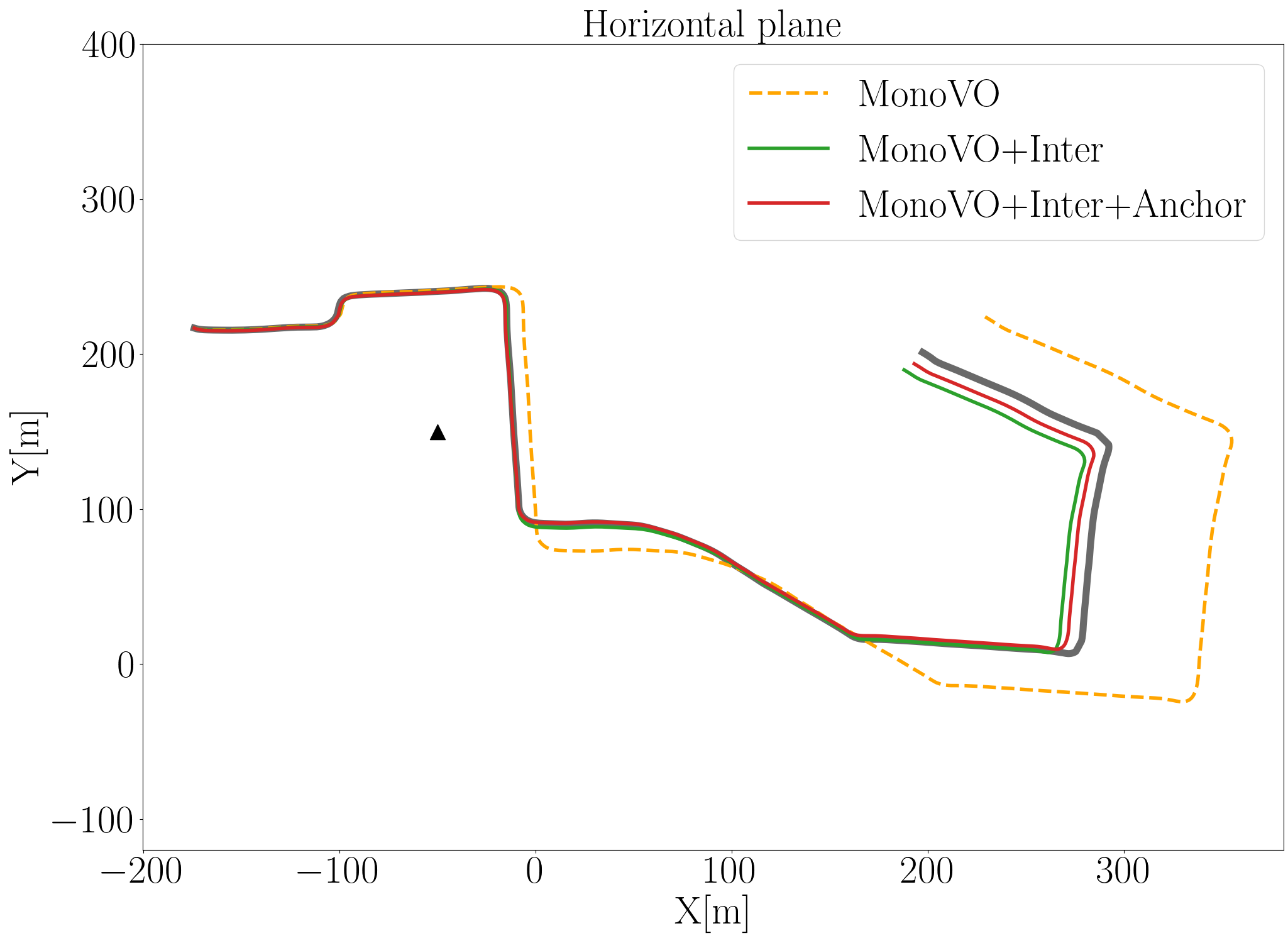}\label{subfig:multiEval_traj_est_agent2_kitti_00}} \\
\subfloat[Rover-3]{\includegraphics[width=.48\linewidth]{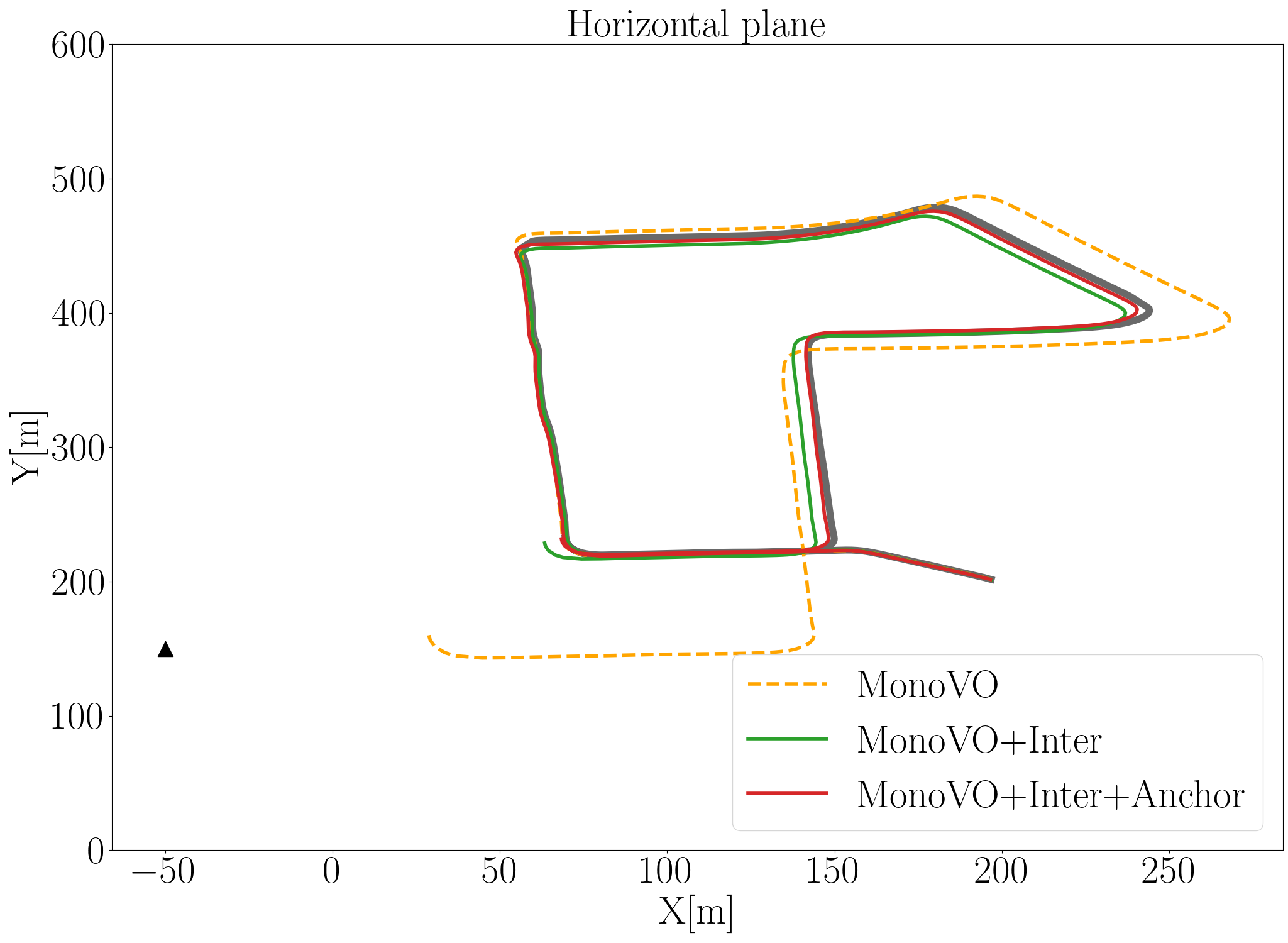}\label{subfig:multiEval_traj_est_agent3_kitti_00}} ~
\subfloat[Rover-4]{\includegraphics[width=.48\linewidth]{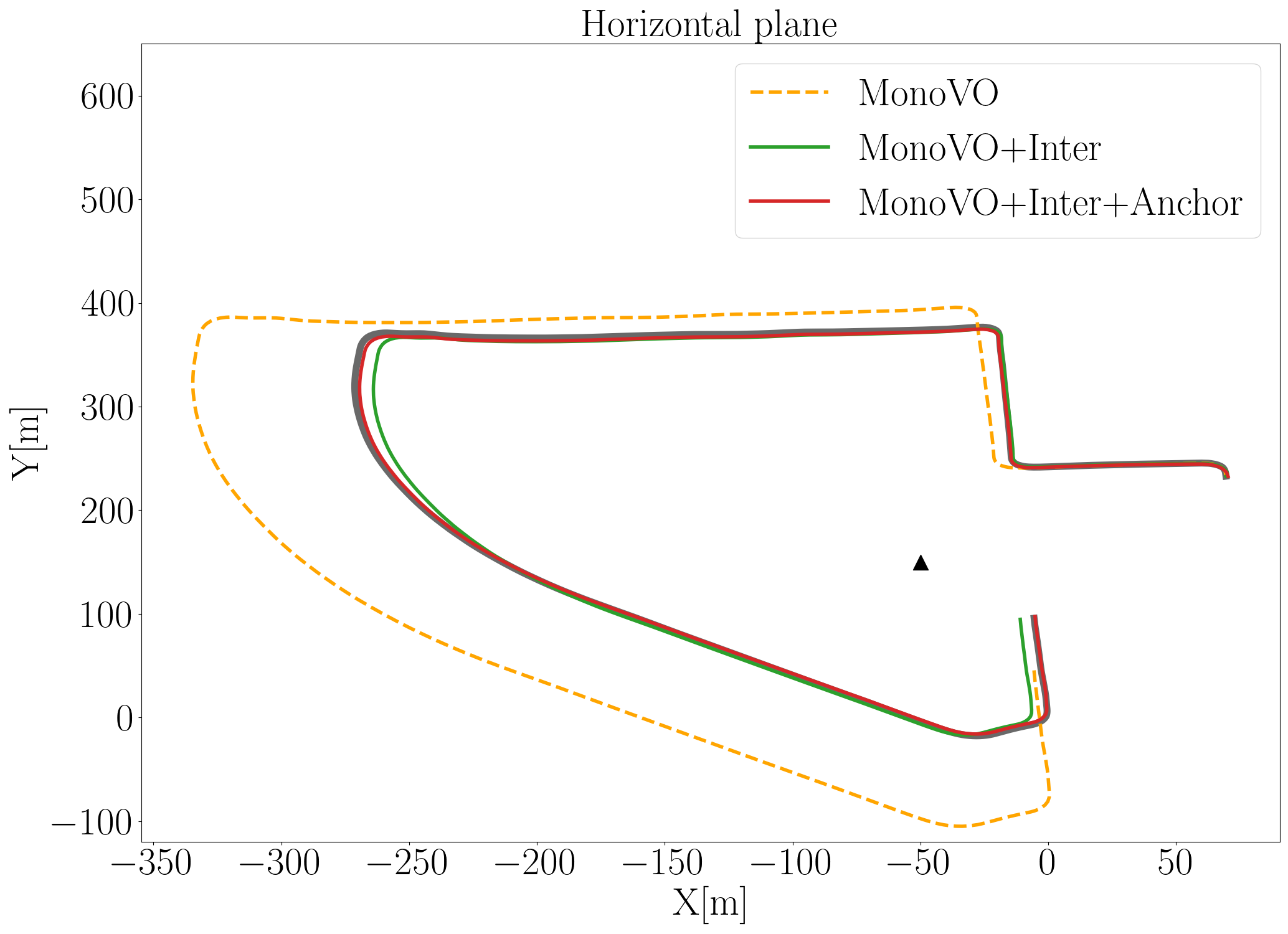}\label{subfig:multiEval_traj_est_agent4_kitti_00}}
\caption{Four agents' horizontal trajectories estimated using only monocular VO (orange), monocular VO and inter-agent ranges (green), and monocular VO, inter-agent ranges, and agent-to-anchor ranges (red). The ground truth trajectories are depicted in gray. The small triangle illustrates the anchor station position.}
\label{fig:multiEval_traj_est_kitti_00} 
\end{figure}

We first analyzed the impact of inter-agent ranges on the positioning accuracy. \figref{fig:multiEval_traj_est_kitti_00} shows the trajectories estimated using only monocular VO (orange), and the ones estimated using monocular VO and inter-agent ranges (without anchor-to-agent connections) are green lines. The ground truth trajectories are shown as gray lines. 
As shown in the figure, the positioning errors of monocular VO were substantially reduced by fusing inter-agent ranges, even when the agents are connected less than $50\%$ of the entire traveling time, as summarized in \figref{fig:multiEval_summary_rangeAvail_kitti_00}. The trajectories estimated using CoVOR-SLAM also follow the ground truth without manual scaling them with the ground truth values. 

Moreover, the trajectories were estimated using CoVOR-SLAM exploiting all available ranges, i.e. both inter-agent and anchor-to-agent ranges (red lines in \figref{fig:multiEval_traj_est_kitti_00}). As shown in the figure, the trajectory estimation errors were further reduced even when the agents were connected only to a single anchor. Following the communication range constraint, the anchor-to-agent ranging was only available when the agent got close enough to the anchor, as shown in \figref{fig:multiEval_summary_rangeAvail_kitti_00}. 

\begin{figure}[t!]
   \centering
   \includegraphics[width=0.95\linewidth]{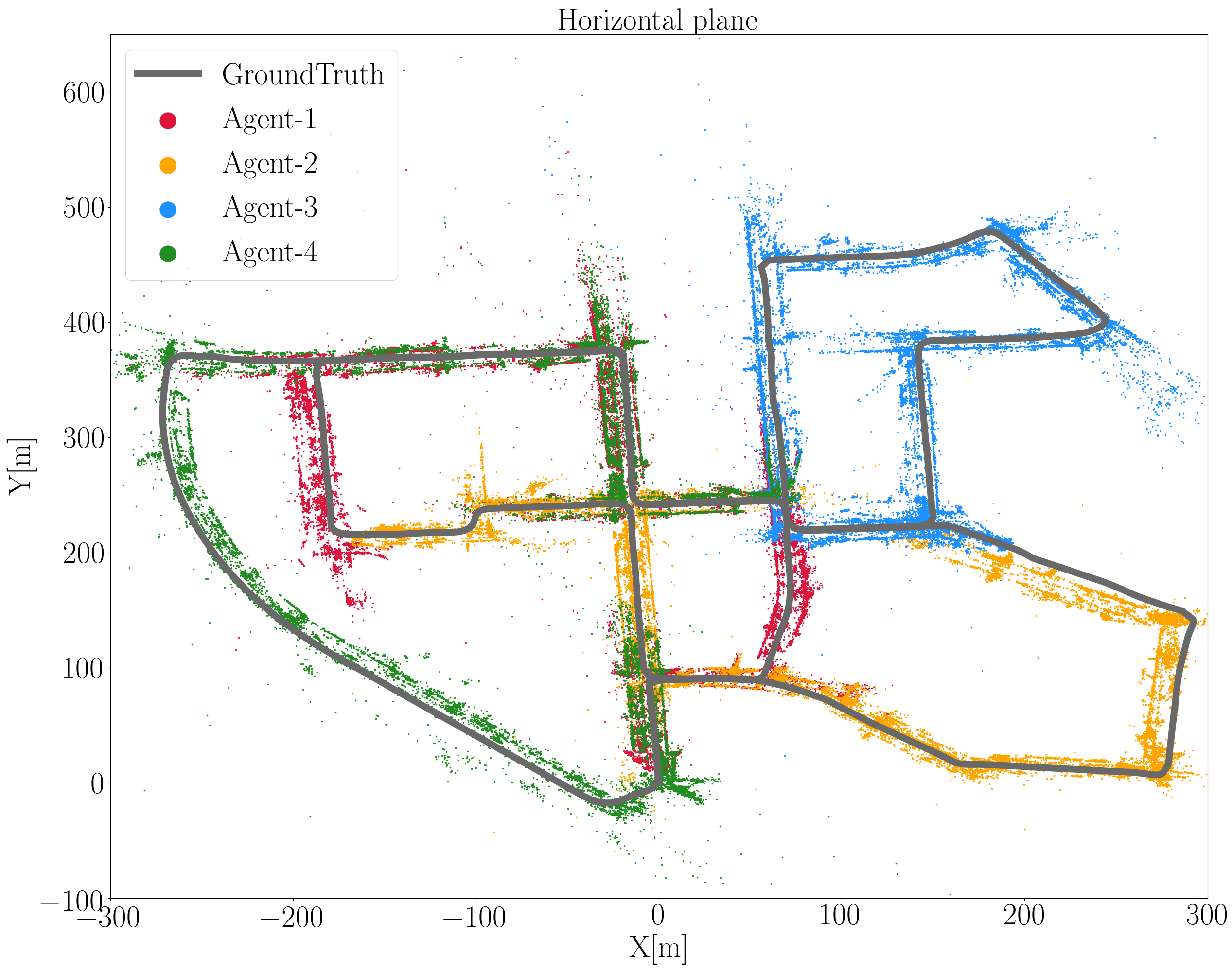}
   \caption{Map points estimated employing four agents using CoVOR-SLAM} 
   \label{fig:multiEval_mp_est_kitti_00}
\end{figure}

\figref{fig:multiEval_mp_est_kitti_00} shows that map points were also accurately estimated using CoVOR-SLAM. Map points should be generated along the ground truth trajectory since the KITTI-00 images were obtained using a front-facing camera on a car in a residential area. As a result, the map points shown in \figref{fig:multiEval_mp_est_kitti_00} can be considered as accurate estimates. 

\subsection{Analysis of the communication loads}
\begin{figure}[t!]
   \centering
   \includegraphics[width=.95\linewidth]{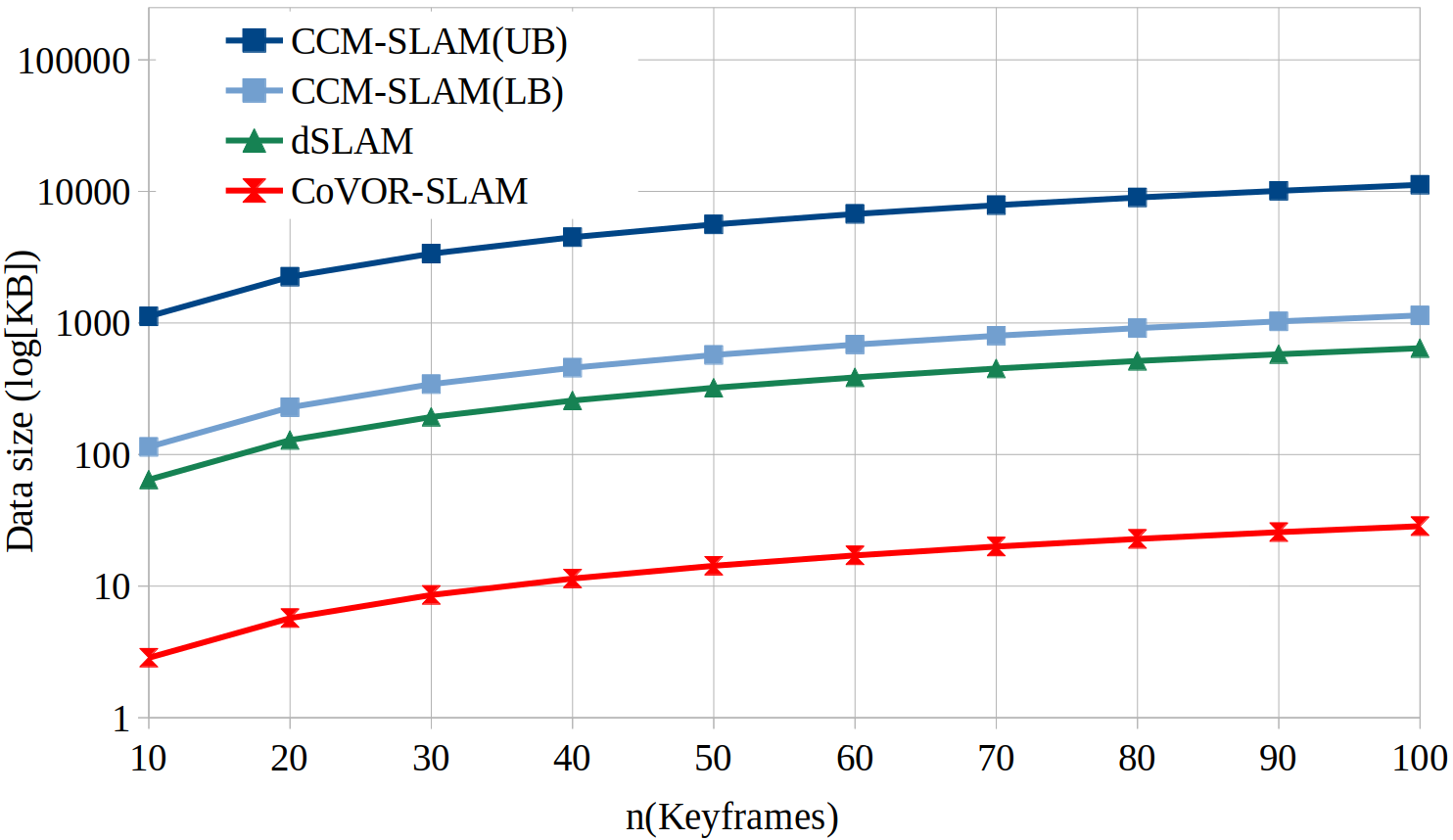}
   \caption{The data size transmitted for collaborative VSLAM with inter-agent loop closures: centralized CCM-SLAM (UB in dark blue and LB in sky blue) and decentralized dSLAM (green). The data size transmitted for CoVOR-SLAM is red} 
   \label{fig:multiEval_data_transmission}
 \end{figure}

The communication requirements of CoVOR-SLAM was compared with cooperative SLAM using inter-agent loop closures. \figref{fig:multiEval_data_transmission} shows the size of data (in a logarithmic scale) transmitted with different numbers of keyframes to run CoVOR-SLAM and two state-of-the-art cooperative SLAM using inter-agent loop closures: centralized CCM-SLAM \cite{schmuck2019ccm} (Upper Bound (UB) in dark blue, and Lower Bound (LB) in sky blue), decentralized dSLAM \cite{cieslewski2018data} (green), and CoVOR-SLAM (red).

As can be seen in the figure, CCM-SLAM requires the greatest communication capability since all agents transmit their local map data to the server and the server carries out the computationally demanding tasks of inter-agent loop closures, such as loop detection, map fusion, and global map optimization. In the dSLAM framework, robots need to transmit about $44\%$ fewer data than the CCM-SLAM's lower bound, which enables collaborative VSLAM running in a decentralized manner.
CoVOR-SLAM further reduces the data size by about $95\%$ than dSLAM, by completely discarding visual information from the keyframe messages. The numbers in the comparison are based on this 4-rover setup. As the size of the swarm increases, the difference in communication load becomes even more significant.
Consequently, CoVOR-SLAM is very useful in environments where communication network capabilities are not sufficient for robots to exchange visual data, such as deep sea and extraterrestrial areas. Moreover, a powerful server is not needed to run CoVOR-SLAM, so the system architecture can be chosen flexibly (centralized or decentralized), depending on the mission and environment. 

\section{Conclusion}
\label{sec:conclusion}

We proposed an approach of cooperative SLAM which complements visual odometry by range measurements (CoVOR-SLAM). The algorithm enables multi-robot systems to accurately estimate poses without requiring significant computing power and communication loads. 
CoVOR-SLAM also solves the scale ambiguity for monocular cameras. 
We tested the CoVOR-SLAM method using experimental platforms, which confirmed all expectations. The same applies to the simulation-based validation with a larger swarm setup with realistic range measurements. The evaluation results also show that the approach is robust with respect to the substantial unavailability of ranging links.


\section*{Acknowledgments}
This work was a part of the VaMEx-CoSMiC project which was supported by the Federal Ministry for Economic Affairs and Energy of the German Bundestag, and administered by the DLR Space Administration [grant number 50NA1521].

\bibliographystyle{IEEEtran}
\bibliography{bibliography}



\begin{IEEEbiographynophoto}{Young-Hee Lee} is a researcher at the Institute of Communications and Navigation of the German Aerospace Center (DLR), Germany, and also a teaching assistant at the Institute for Communications and Navigation of Technical University of Munich, Germany. She received her BS in Aerospace Engineering from Korea Advanced Institute of Science and Technology (KAIST), South Korea, and her MS also from KAIST for her research on satellite optimal control. She is currently pursuing her Ph.D. on cooperative localization for a multi-agent system in communication-constrained environments using vision sensors and inter-agent range networks. 
\end{IEEEbiographynophoto}

\begin{IEEEbiographynophoto}{Chen Zhu} is a senior research fellow and the head of Visual and Terrestrial Augmentation Group at the Institute of Communications and Navigation, German Aerospace Center (DLR). He received his Ph.D. degree (Dr.-Ing.) and Master’s degree (M.Sc.) from Technical University of Munich, Germany and his Bachelor’s degree from Tsinghua University, Beijing, China. He is interested in the research fields of visual navigation, multi-sensor fusion, and robotic swarm navigation, currently focusing on the system integrity.
\end{IEEEbiographynophoto}

\begin{IEEEbiographynophoto}{Thomas Wiedemann} obtained a Bachelor and Master degree from the Faculty of Mechanical Engineering at the Technical University of Munich. Since 2014, he is a scientist at the Institute of Communications and Navigation at German Aerospace Center (DLR), where he is working in the Swarm Exploration Research Group. From 2017 to 2020, he was enrolled as a PhD student at the Centre for Applied Autonomous Sensor Systems (AASS) at \"{O}rebro University in Sweden, where he received his PhD in computer science in 2020 on the topic of Domain Knowledge Assisted Robotic Exploration and Source Localization. 
\end{IEEEbiographynophoto}

\begin{IEEEbiographynophoto}{Emanuel Staudinger} received the {M.Sc.} degree in Embedded Systems Design from the University of Applied Sciences of Hagenberg, Austria, in 2010. Since 2010, he is with the Institute of Communications and Navigation of the German Aerospace Center (DLR), Wessling, Germany. 
He received a {Ph.D.} with distinction from the Institute of Electrodynamics and Microelectronics at the University of Bremen, Germany, in 2015. His current research interests include system design for cooperative positioning, experimental platform design based on SDRs, and experimental validation for swarm navigation. 
\end{IEEEbiographynophoto}

\begin{IEEEbiographynophoto}{Siwei Zhang} (Member of IEEE) received his B.Sc. in electrical engineering from Zhejiang University, China, in 2009, his M.Sc. in communication engineering from the Technical University of Munich, Germany, in 2011, and his Dr.-Ing. (Ph.D.) in electrical engineering from the University of Kiel, Germany, in 2020. Since 2012, he has been a researcher at the Institute of Communications and Navigation of the German Aerospace Center (DLR). His research interests lie in statistical signal processing in wireless communication and navigation, particularly in multi-agent joint communication, navigation and sensing. 
He is a recipient of the 2021 DLR Science Award and several Best Paper/Presentation Awards at international conferences including 2015 and 2022 ION GNSS+, 2021 IEEE CCNC-RoboCom and 2023 IEEE Aerospace Conference. Siwei Zhang is an external lecturer at Technical University of Munich and University of Kiel for robotic swarm communications and navigation since 2023.
\end{IEEEbiographynophoto}

\begin{IEEEbiographynophoto}{Christoph G\"{u}nther}
Christoph G\"{u}nther studied theoretical physics at the Swiss Federal Institute of Technology in Zurich, Switzerland. From 1989 to 2003 he worked in industry at Asea Brown Boveri, Ascom and Ericsson on research in cryptography, coding, communications, and information theory, as well as in the development of mobile phones. From 2003 until 2023 he was the Director of the Institute of Communication and Navigation at the German Aerospace Center (DLR) and since 2004 additionally a full professor at TU-M\"{u}nchen (TUM). He retired lately from both positions.
\end{IEEEbiographynophoto}

 
%




\end{document}